\documentclass[twocolumn]{svjour3}

\smartqed  % flush right qed marks, e.g. at end of proof
\usepackage{amssymb,amsmath}
\usepackage{microtype}
\usepackage{graphicx}
\usepackage{booktabs}
\usepackage{multirow}
\usepackage[usenames,dvipsnames,svgnames,table]{xcolor}
\usepackage[pagebackref=true, breaklinks=true, colorlinks, bookmarks=false]{hyperref}
\usepackage{subfig}
\usepackage{cite}
\usepackage{nicefrac}

\newcommand{\x}{\mathbf{x}}
\newcommand{\X}{\mathbf{X}}

\newcommand{\w}{\mathbf{w}}

\newcommand{\WM}{\mathbf{W}}
\newcommand{\G}{\mathbf{\Gamma}}

\DeclareMathOperator{\sgn}{sgn}
\DeclareMathOperator{\sigmoid}{sigmoid}

\journalname{IJCV}

\begin{document}

\title{Deep Hashing with Hash-Consistent Large Margin Proxy Embeddings}

\author{Pedro Morgado \and Yunsheng Li \and Jose Costa Pereira \and Mohammad Saberian \and Nuno Vasconcelos}


\institute{Pedro Morgado\textsuperscript{*}, Yunsheng Li and Nuno Vasconcelos \at
           Department of Electrical and Computer Engineering\\
           University of California, San Diego \\
           \textsuperscript{*}\email{pmaravil@eng.ucsd.edu}
           \and
           Jose Costa Pereira \at
           Huawei Technologies
           \and
           Mohammad Saberian \at
           Netflix
}

\date{Received: date / Accepted: date}

\maketitle

\begin{abstract}
  Image hash codes are produced by binarizing the embeddings of convolutional
  neural networks (CNN) trained for either classification or
  retrieval. While proxy embeddings achieve good performance on both tasks,
  they are non-trivial to binarize,
  due to a rotational ambiguity that encourages non-binary embeddings.
  The use of a fixed set of proxies (weights of the CNN classification layer)
  is proposed to eliminate this ambiguity, and a procedure to design
  proxy sets that are nearly optimal for both classification and hashing
  is introduced. The resulting {\it hash-consistent
    large margin\/} (HCLM) proxies are shown to encourage saturation of
  hashing units, thus guaranteeing a small binarization error, while
  producing highly discriminative hash-codes. A semantic extension (sHCLM),
  aimed to improve hashing performance in a transfer scenario, is also
  proposed. Extensive experiments show that sHCLM embeddings achieve
  significant improvements over state-of-the-art hashing procedures on
  several small and large datasets, both within and beyond the set of
  training classes.
\end{abstract}

\section{Introduction}

% Retrieval & hashing.
Image retrieval is a classic problem in computer vision. 
Given a query image, a nearest-neighbor search is performed on an
image database, using a suitable image representation and similarity
function~\cite{Smeulders:PAMI00}. Hashing methods enable efficient
search by representing each image with a binary string,
known as the \textit{hash code}.  This enables efficient indexing
mechanisms, such as hash tables, or similarity functions, such as Hamming
distances, implementable with logical operations.  
The goal is thus to guarantee that similar images are represented
by similar hash codes~\cite{Andoni:FOCS06,Datar:LSH,Mu:CAI10}.

\begin{figure*}[t!]
  \centering
  \includegraphics[width=\linewidth]{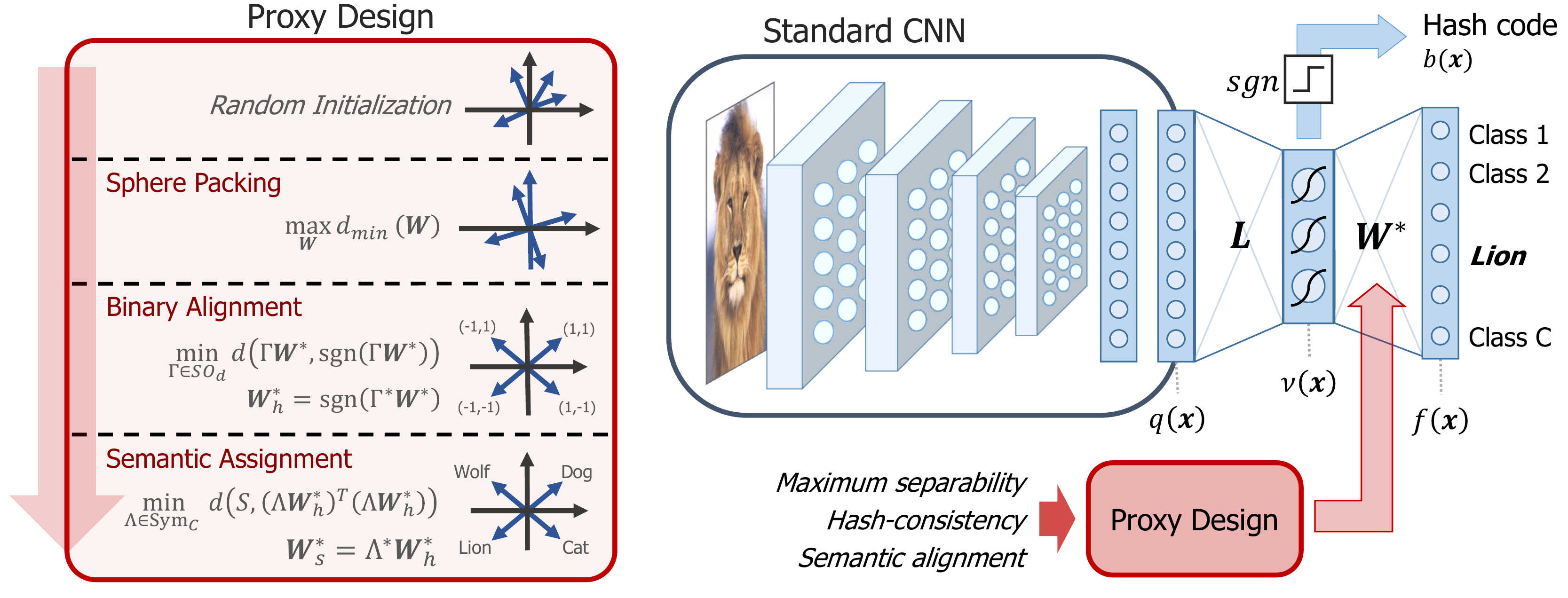}
	
  \caption{Deep hashing CNN with sHCLM proxies. Left: Proxy design.
    A set of large margin proxies is first generated by solving
    a sphere packing problem. A rotation is next applied to render these
    proxies as binary as possible, and their entries binarized to produce the
    HCLM proxy set ${\bf W}_h^*$. Finally, a semantic assignment maps
    proxies to classes, producing the sHCLM proxy set ${\bf W}_s^*$
    used as weight matrix ${\bf W}^*$ of the CNN.
    Right: Deep hashing CNN.  The output of layer $\nu(\x)$ is a
    hashing function, binarized to produce a bit-string $b(\x)$ for fast
    image retrieval.}
	\label{fig:deephash}
\end{figure*}

% Modern hashing approaches: Description
Early hashing techniques approximated nearest neighbor searches between
low-level features~\cite{Datar:LSH,Mu:CAI10,Gong:ITQ,Weiss:SH}.
However, humans judge similarity based on image \textit{semantics}, such
as scenes, objects, and attributes. 
This inspired the use of semantic representations for image
retrieval~\cite{Lampert:Attributes,Nikhil:QBSE,Li:ObjectBank} and, by
extension, hashing~\cite{Xia:CNNH,Lin:BHC,Zhang:BitScalable,Zhong:DHN}. 
Modern hashing techniques rely on semantic embeddings implemented with
convolutional neural networks (CNNs), as illustrated in
Figure~\ref{fig:deephash} (right). A CNN feature extractor $q(\x)$ is
augmented with a hashing layer that outputs a nearly-binary
code $\nu(\x)$ using saturating non-linearities, such as a $\sigmoid$
or $\tanh$. The code $\nu(\x)$ is thresholded to produce a
bitstream $b(\x)$, which is the hash code for image $\x$. 

The embedding $\nu(\x)$ can be learned by
metric learning~\cite{Wang:TripletSH,Lai:NINH,liu2016deep} or
classification~\cite{Lin:BHC,Yang:SSDH,Li:DeepSDH}, with
classification methods usually being preferred for recognition and
metric learning for retrieval. 
However, it has recently been shown that good retrieval performance
can also be achieved with {\it proxy embeddings\/}~\cite{movshovitz2017}
derived from the neighborhood component analysis (NCA)~\cite{goldberger:NCA}
metric learning approach.
These embeddings learn a set of {\it proxies,\/} or class
representatives, around which class examples cluster. 
Learning involves minimizing a variant of the softmax loss defined
by a pre-chosen distance function. For standard distance functions,
proxies are identical to the columns ${\bf w}_y$ of the weight matrix
$\bf W$ of the softmax layer of the classifier of Figure~\ref{fig:deephash}. 
Under such distances, there is little difference between a
classifier and a proxy embedding. The architecture of
Figure~\ref{fig:deephash} can thus be used both for both classification
or retrieval.

% Binarization - the problem
Nevertheless, generic embeddings are unsuitable for hashing, where the
outputs of $\nu(\x)$ should be binary.
The goal is to make the $\sigmoid$ or $\tanh$ non-linearities
of Figure~\ref{fig:deephash} saturate without degrading classification or
retrieval performance. This is difficult because
classification and metric learning losses are invariant to rotation. 
For example, classification losses only depend on the dot-products
of $\nu(\x)$ and the rows $\w_y$ of the matrix $\bf W$. 
Hence, even if there is a solution $({\bf W}, \nu(\x))$ with binary
$\nu(\x)$, an infinite number of \textit{non-binary solutions of equivalent
  loss} can be constructed by rotating both the proxies $\w_y$ and the
embedding $\nu(\x)$.  In the absence of further constraints, there is no
incentive to learn a binary embedding.
Hashing techniques address this by the addition of loss terms
that encourage
a binary $\nu(\x)$~\cite{Liong:SDH,Shen:SDH,Yang:SSDH,Lin:UnsHashing,Cao:DQN}. 
In general, however, this degrades both classification and retrieval
performance.

We address this problem by leveraging the observation that the optimally
discriminant embedding $\nu(\x)$ of image $\x$ must be as aligned as
possible (in the dot-product sense) with the proxy $\w_y$ of the
corresponding class $y$. The rotational ambiguity can thus be removed by
using a \textit{fixed} set of weights $\bf W$ and learning the optimal
embedding $\nu(\x)$ {\it given\/} these fixed proxies. This allows
the encoding in the proxy set $\bf W$ of any properties desired for $\nu(\x)$.
In this work, we consider the design of proxy sets ${\bf W}^*$ that are
nearly optimal for \textit{both} classification and hashing. 
This involves two complementary goals. On the one hand, classification
optimality requires \textit{maximum separation} between proxies. 
On the other hand, hashing optimality requires {\it binary\/} proxies.

We show that the first goal is guaranteed by any rotation of the solution of
a classical sphere packing optimization problem, known as the Tammes
problem~\cite{Tammes1930}. Drawing inspiration from the classical iterative
quantization (ITQ) procedure of \cite{Gong:ITQ}, we then seek the rotation
of the Tammes solution that makes these proxies most binary.
This produces a set of \textit{hash-consistent large-margin} (HCLM) proxies.
Unlike ITQ, which rotates
the embedding $\nu(\x)$, the proposed \textit{binary alignment} is
applied to the proxies only, i.e. before $\nu(\x)$ is even learned.
The embedding can then be learned end-to-end, guaranteeing that
it is optimally discriminant. Also, because this requires 
$\nu(\x)$ to be aligned with the proxies, training $\nu(\x)$ for
classification \textit{also} forces its outputs to saturate, eliminating
the need for additional binarization constraints. Finally, because the
proxies are not learned, learning is freed from rotational ambiguities.

Beyond rotations, the Tammes solution is also invariant to proxy permutations.
We leverage this additional degree of freedom to seek the proxy-class
assignments that induce a semantically structured $\nu(\x)$, where
similar proxies represent similar classes. This is denoted
the {\it semantic HCLM\/} (sHCLM) proxy set.
Since semantically structured embeddings enable more effective transfer
to unseen classes~\cite{Lampert:Attributes,Akata:ALE,Morgado:SCoRe}, 
this enhances retrieval performance in transfer
scenarios~\cite{Sablayrolles:SHEval}.
The steps required for the generation of an sHCLM proxy set are summarized
on the left of Figure~\ref{fig:deephash}. 

Extensive experiments show that sHCLM proxy embeddings achieve
state-of-the-art hashing results on several small and large scale datasets,
for both classification and retrieval, both within and beyond the set of
training classes. We also investigate the combination of proxy and classical
triplet embeddings. This shows that their combination is unnecessary
for datasets explicitly annotated with classes but can be useful for
multi-labeled datasets, where the class structure is only defined implicitly
through tag vectors.

\section{Related Work}
In this section, we review previous work on image embeddings, retrieval, hashing and transfer learning techniques that contextualize our contributions.

\paragraph{Image retrieval:}
Content-based image retrieval (CBIR) aims to retrieve images from large databases based on their visual content alone. Early systems relied on similarities between low-level image properties such as color and texture~\cite{Flickner:QBIC,bach1996virage,Smith:VisualSEEk}. However, due to the semantic gap between these low-level image representations and those used by humans, such systems had weak performance~\cite{Smeulders:PAMI00}. This gap motivated substantial research in semantic image embeddings that better align with human judgments of similarity. Early works include query by semantic example~\cite{Nikhil:QBSE}, semantic multinomials~\cite{pereira2014cross}, classeme representations~\cite{torresani2010efficient}, and object banks~\cite{Li:ObjectBank}. These methods used embeddings
learned by generative models for images or binary classifiers, usually support vector machines. More recently, CNNs have been used to extract more robust semantic embeddings, with improved retrieval performance~\cite{gordo2016deep}.

 \paragraph{Embeddings:}
 Many algorithms have been proposed to learn embeddings $\nu({\bf x})$ endowed
 with a metric, usually the Euclidean distance. For this, pairs of examples
 in a dataset are labeled  ``similar'' or ``non-similar,'' and a CNN
is trained with a loss function based on distances between pairs or
triplets of similar and non-similar examples, e.g.,
the {\it contrastive loss\/} of~\cite{hadsell2006} (pairs) or
the {\it triplet loss\/} of~\cite{weinberger2009}. 
These embeddings have been successfully applied to object retrieval~\cite{bell2015}, face verification~\cite{sun2014,schroff2015}, image retrieval~\cite{wang2014}, clustering~\cite{oh2016}, person re-identification~\cite{zhang2019scan,zhang2017image} among other applications.
% A difficulty of these approaches is their learning complexity, due to the fact that ther are exponentially more pairs or triplets than examples in any given dataset.
% While sampling strategies~\cite{schroff2015, wang2014, oh2016, sun2014} attenuate this problem, it is often impossible to sample a substantial percentage of the dataset pairs or triplets. Consequently, these approaches converge slowly and are difficult to train. 

%Faster convergence can be achieved by combining metric learning
%and classification, complementing the embedding loss with a class-based
%cross-entropy loss~\cite{bell2015, sun2014}. This is particularly
%beneficial when the number of classes is large~\cite{sun2014}.
Another possibility is to define a class-based embedding. This is rooted in the
neighborhood component analysis (NCA) procedure~\cite{goldberger:NCA}, based
on a softmax-like function over example distances. However, because NCA
requires a normalization over the entire dataset, it can be intractable.
Several approximations replace training
examples by a set of proxies. \cite{sohn2016} proposed the
{\it N-tuplet loss,\/} which normalizes over (N+1)-tuplets of examples
and \cite{wen2016} introduced the {\it center loss,\/} which combines a
softmax classifier and an additional term defined by class centers.
Finally, \cite{movshovitz2017} replaces the $N$-tuplet of examples by a set
of learned proxies, or class representatives, around which examples
cluster. \cite{movshovitz2017} showed that proxy embeddings outperform
triplet embeddings~\cite{schroff2015,oh2016}, $N$-tuplet
embeddings~\cite{sohn2016}, and the method of \cite{song2} on various
retrieval tasks. 

\paragraph{Hashing:}
Computational efficiency is a major concern for retrieval,
since nearest-neighbor search scales poorly with database size.
Hashing techniques, based on binary embeddings, enable 
fast distance computations using Hamming distances. This has made
hashing a popular solution for retrieval.
Unsurprisingly, the hashing literature experienced a trajectory similar to
image retrieval.
Early approaches were unsupervised~\cite{Datar:LSH, Mu:CAI10, Gong:ITQ, Weiss:SH}, approximating nearest-neighbor search in Euclidean space using fast bit
operations. Semantic supervision was then introduced to fit the human
notion of similarity~\cite{Wang:SSH,Kulis:BRE,Norouzi:MLH,Liu:KSH}.
This was initially based on handcrafted features, which limit retrieval
performance. More recently, CNNs became dominant. In one of the earliest
solutions, PCA and discriminative dimensionality reduction of CNN
activations were used to obtain short binary
codes~\cite{Babenko:NeuralCodes}. More commonly, the problem is
framed as one of joint learning of hash codes and semantic features, with
several approaches proposed to achieve this goal, e.g., by exploiting
pairwise similarities~\cite{Xia:CNNH,Liong:SDH,Zhu:DHN,Cao:DQN,Li:DeepSDH,
  jiang2018asymmetric}, triplet
losses~\cite{Lai:NINH,Zhang:BitScalable,Wang:TripletSH,Zhang:QaDWH} or
class supervision~\cite{Xia:CNNH,Yang:SSDH,Lin:BHC,Zhong:DHN,
  Li:DeepSDH,Zhang:QaDWH}.

As shown in Figure~\ref{fig:deephash}, most approaches introduce a layer of squashing non-linearities (e.g., $\tanh$ or $\sigmoid$) that, when saturated, produces a binary code. Extensive research has been devoted to the regularization of these networks with losses that favor saturation, using constraints such as maximum entropy~\cite{Liong:SDH,Yang:SSDH,Lin:UnsHashing,Huang:UTH}, independent bits~\cite{Liong:SDH}, low quantization loss~\cite{Liong:SDH,Shen:SDH,Yang:SSDH,Lin:UnsHashing,Cao:DQN}, rotation invariance~\cite{Lin:UnsHashing,Huang:UTH}, low-level code consistency~\cite{Zhang:BitScalable}, or bimodal Laplacian quantization priors~\cite{Zhu:DHN}. These constraints compensate for the different goals of classification and hashing, and are critical to the success of most hashing methods. However, they reduce the discriminant power of the CNN. Like these methods, the proposed hashing procedure leverages the robust semantic embeddings produced by CNNs. However, the proposed CNN architecture meets the binarization
requirement without the need for binarization losses. We show that, for CNNs
with sHCLM proxies,  the cross-entropy loss suffices to guarantee
binarization.

\paragraph{Transfer protocol for supervised hashing:}
Traditional hashing evaluation protocols are class-based. A retrieved image
is relevant if it has the same class label as the query or at least one
label for multi-labeled datasets. However, a classifier that outputs
a single bit per class enables high retrieval performance with extremely small
hash codes \cite{Sablayrolles:SHEval}. The problem is that the traditional
protocol ignores the fact
that retrieval systems deployed in the wild are frequently confronted with
images of classes unseen during training. Ideally, hashing should generalize
to such classes. This is unlikely under the traditional protocol, which
encourages hash codes that overfit to training classes.

To avoid this problem, \cite{Sablayrolles:SHEval} proposed  to measure
retrieval performance on images from previously unseen classes. While some
recent works have adopted this
protocol~\cite{Sablayrolles:SHEval,jain2017subic,lu2017deep}, they
still disregard transfer during CNN training.
We propose a solution to this problem by explicitly encoding the semantic
structure in the sHCLM proxies used for training. Our approach is inspired
by the use of semantic spaces for zero-shot and few-shot learning
problems~\cite{Akata:ALE, Morgado:SCoRe}, such as those induced by
attributes~\cite{Lampert:Attributes}, large text
corpora~\cite{frome2013devise} or other measures of class
similarity~\cite{rohrbach2011evaluating}. However, because the goal is to
improve hashing performance, proxies need to be optimal for hashing as well.
To accomplish this, we propose a procedure to semantically align
hash-consistent proxies, with apriori measures of class similarity.

\section{Hashing with the Proxy Embedding}
\label{sec:limitations}

Modern hashing algorithms are implemented with CNNs trained for either classification or metric learning. A CNN implements an embedding $\nu: {\cal X} \rightarrow {\cal V} \subset \mathbb{R}^d$ that maps image $\x \in {\cal X}$ into a $d$-dimensional feature vector ${\bf v} = \nu(\x)$. In this section, we discuss the limitations of current embeddings for hashing.

\subsection{Classification vs. Metric Learning}

For classification, image $\x$ belongs to a class
drawn from random variable $Y \in \{1, \ldots, C\}$.
Given a dataset ${\cal D} = \{(\x_i,y_i)\}_{i=1}^n$, a CNN is
trained to discriminate the $C$ classes by minimizing the empirical
cross-entropy loss
\begin{equation}
  {\cal R} = - \sum_i \log P_{Y|\X}(y_i|\x_i),
  \label{eq:xentrisk}
\end{equation}
where
$P_{Y|\X}(y|\x)$ are posterior class probabilities modeled by softmax regression
\begin{equation}
  P_{Y|\X}(y|\x) 
  = \frac{e^{\w_y^T \nu(\x) + b_y}}{\sum_k e^{\w_k^T \nu(\x) + b_k}},
  \label{eq:softmax}
\end{equation}
where $\w_y$ is the parameter vector for class $y$ and $b_y$ the class bias. 

For metric learning, the goal is to endow the feature space $\cal V$ with a
metric, usually the squared Euclidean distance, to allow operations like
retrieval. Although seemingly different, metric learning and classification
are closely related.  As shown in Appendix~\ref{app:equivalence}, learning
the embedding $\nu(\x)$ with~(\ref{eq:xentrisk}) using the softmax classifier
of~(\ref{eq:softmax}) is equivalent to using
\begin{equation}
  P_{Y|\X}(y|\x) = \frac{e^{-d_\phi(\nu(\x),\mu_y)}}
  {\sum_{k} e^{-d_\phi(\nu(\x),\mu_k)}},
\end{equation}
where $d_\phi$ is a Bregman divergence~\cite{bregman1967} and $\mu_y$ the mean
of class $y$. Hence, any classifier endows the embedding $\nu(\x)$
with a metric $d_\phi$. This metric is the Euclidean distance
if and only if
\begin{equation}
  P_{Y|\X}(y|\x) = \frac{e^{-d_\phi(\nu(\x),\w_y)}}
  {\sum_{k} e^{-d_\phi(\nu(\x),\w_k)}},
  \label{eq:sy}
\end{equation}
in which case (\ref{eq:sy}) is identical to
(\ref{eq:softmax}). In sum, training an embedding with the Euclidean
distance is fundamentally not very different from training the softmax
classifier.

In fact, the combination of~(\ref{eq:xentrisk}) and~(\ref{eq:sy}) is
nearly identical to the \textit{proxy embedding} technique
of~\cite{movshovitz2017}. This is a metric learning approach derived from
neighborhood component analysis (NCA)~\cite{goldberger:NCA}, which denotes
the parameters $\w_y$ as a set of proxy vectors. 
The only difference is that, in NCA, the probabilities of (\ref{eq:sy}) are not properly normalized, since the summation in the denominator is taken over $k \neq y$. However, because there are usually many terms in the summation, the practical difference is small.
Therefore, we refer to an embedding $\nu(\x)$ learned by cross-entropy minimization with either the model of~(\ref{eq:softmax}) or~(\ref{eq:sy}) as the {\it proxy embedding.\/} 
This is a unified procedure for classification and metric learning, where the parameters $\w_y$ can be interpreted as either classifier parameters or metric learning proxies.

\subsection{Deep Hashing}
\label{sec:deep_hashing}

Given an embedding $\nu(\x)$ endowed with a metric $d(\cdot,\cdot)$, image retrieval can be implemented by a nearest-neighbor rule. The query $\bf x$ and database images ${\bf z}_i$ are forward through the CNN to obtain the respective vector representations $\nu(\x)$ and $\nu({\bf z}_i)$, and database vectors $\nu({\bf z}_i)$ are ranked by their similarity to the query $\nu(\x)$. This requires floating point arithmetic for the metric $d(\cdot,\cdot)$ and floating point storage for the database representations $\nu({\bf z}_i)$, which can be expensive. Hashing aims to replace $\nu(\x)$ with a bit-string $b(\x)$, known as the \textit{hash code}, and $d(\cdot,\cdot)$ with a low complexity metric,  such as the Hamming distance
\begin{equation}
    d(\x,{\bf z}_i) = \sum_c b_c(\x) \oplus b_c({\bf z}_i) 
    \label{eq:hamming}
\end{equation}
where $\oplus$ is the XOR operator.

\begin{figure*}[t!]
	\centering
	\includegraphics[width=0.75\linewidth]{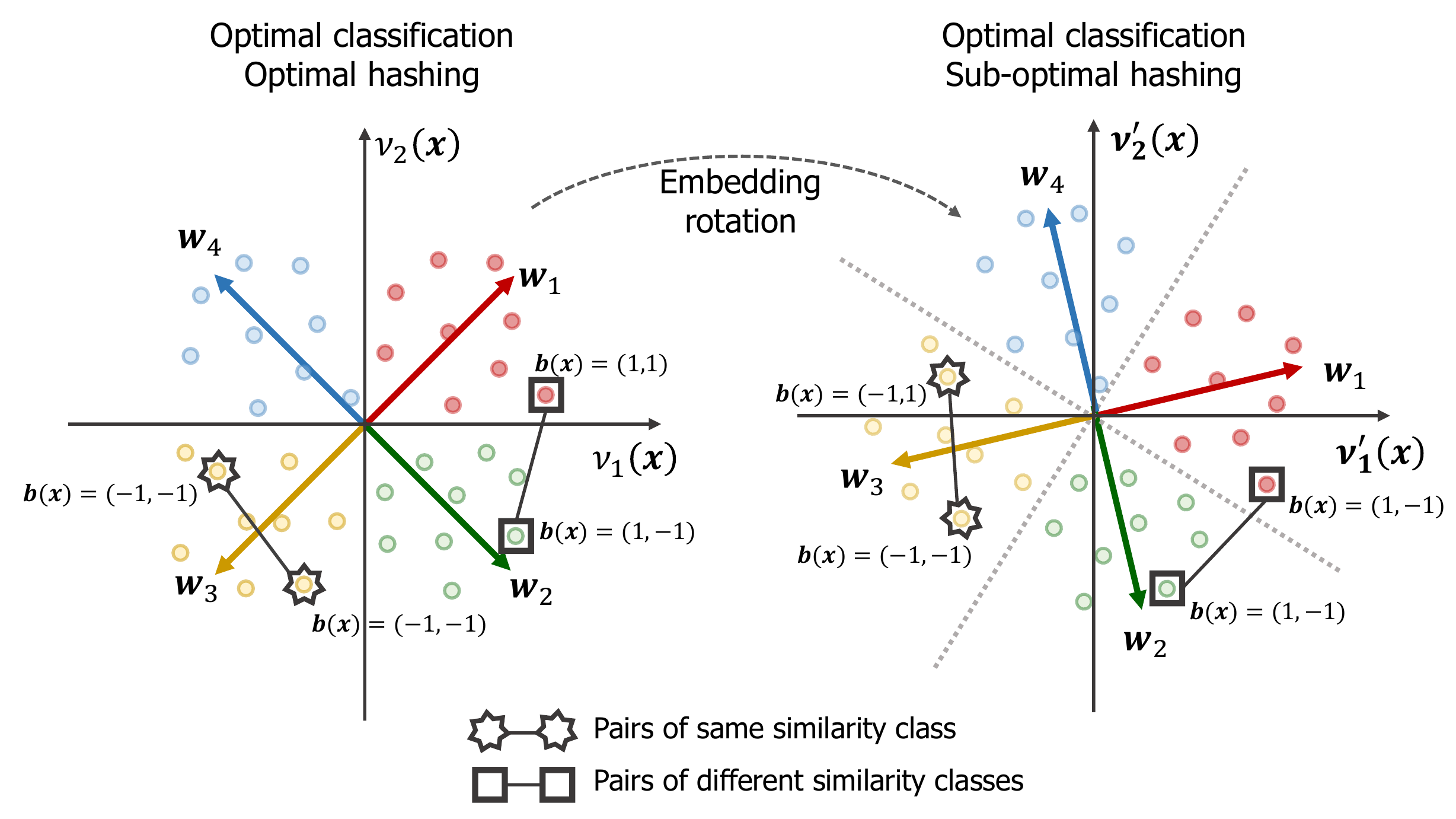}
	\caption{Effects of rotation on optimality for classification and hashing. The figure depicts two possible embeddings for classification or hashing. Because the two solutions differ by a joint rotation of feature vectors $\nu(\x)$ and proxies $\w_y$, and the loss of~\eqref{eq:xent2} only depends on the dot-products $\langle \nu(\x),\w_y\rangle$, the two solutions are equally optimal for classification. This is, however, not the case for hashing. While, on the left, binarization maps all the examples from the same class into the same hash code $b(\x)$, this is not true on the right. Best viewed in color.}
	\label{fig:intuition}
\end{figure*}

In the hashing literature, the proxy embedding is frequently used to obtain hash codes. Figure~\ref{fig:deephash} (right) illustrates the architecture commonly used to produce $b(\x)$. A CNN encoder first extracts a feature representation $q(\x)$ from image $\bf x$. This is then mapped into the low-dimensional embedding $\nu(\x) \in \mathbb{R}^d$. A $d$-bit hash code is finally generated by thresholding $\nu(\x)$
\begin{equation}
    b(\x) = \sgn(\nu(\x)),
    \label{eq:b}
\end{equation}
where $\sgn(\cdot)$ is the vector of signs of its entries. 
This network is trained for either classification or metric learning, using a softmax regression layer of the form of~(\ref{eq:softmax}) or~(\ref{eq:sy}), respectively. 
The network parameters are trained to optimize (\ref{eq:xentrisk}). 

For hashing, the binarization error of (\ref{eq:b}) must be as small as possible. This is encouraged by implementing the mapping $q(\x)\to \nu(\x)$ as
\begin{equation}
    \nu(\x) = \tanh({\bf L}^T q(\x)+{\bf b}) \in \mathbb{R}^d,
    \label{eq:f}
\end{equation}
where $\bf L$ is a dimensionality reduction matrix, ${\bf b}$ a bias vector and $\tanh(\cdot)$ an element-wise squashing non-linearity. The introduction of these non-linearities encourages $\nu(\x)$ to be binary by saturation, i.e.~by making the output of $\tanh(\cdot)$ close to its asymptotic values of $+1$ or $-1$. Under this assumption, $b(\x) \approx \nu(\x)$ and there is no information loss due to the binarization of \eqref{eq:b}. Since the cross-entropy risk encourages $\nu(\x)$ to maximally discriminate similarity classes, the same holds for the hash codes $b(\x)$. These properties have made the architecture of Figure~\ref{fig:deephash} popular for hashing~\cite{Xia:CNNH,Yang:SSDH,Lin:BHC,Zhong:DHN,Li:DeepSDH,Zhang:QaDWH}.

\subsection{Challenges}
\label{sec:challenges}

The discussion above assumes that it is possible to obtain a discriminative embedding $\nu(\x)$ with saturated non-linearities, by training the CNN of Figure~\ref{fig:deephash} for classification or metric learning. However, this problem
does not have a unique solution. Even when it is optimal for $\nu(\x)$ to saturate, many equivalent solutions do not exhibit this behavior. This has been experimentally observed by previous works, which proposed learning the CNN with regularization losses that penalize large binarization errors~\cite{Liong:SDH, Shen:SDH, Yang:SSDH, Lin:UnsHashing, Cao:DQN}. In our experience, these approaches fail to guarantee \textit{both} optimal classification and saturation of hash scores. Instead, there is usually a trade-off, where emphasizing one component of the loss weakens performance with respect to the other.

This can be understood by writing the cross-entropy as
\begin{equation}
    L_p(\x, y) = -\log\frac{e^{\langle \nu(\x), \w_y\rangle}}
    {\sum_k e^{\langle \nu(\x), \w_k\rangle}}.
    \label{eq:xent2}
\end{equation}
Since (\ref{eq:xent2}) is minimum when $\left<\nu(\x), \w_y\right>$ is much
larger than all other $\left<\nu(\x), \w_k\right>$, cross-entropy
minimization encourages the embedding to align with the proxy of the
ground-truth class $y$
\begin{equation}
    \nu^*(\x) \propto \w_y.
    \label{eq:hopt}
\end{equation}
This is illustrated in Figure~\ref{fig:intuition} for a classifier with $C=4$ classes, proxies $\w_c$ and embedding $\nu(\x)$ of dimension $d=2$. The embeddings $\nu(\x)$ of each similarity class (points of a given color) cluster around the corresponding proxy (vector of the same color). However, this solution is not unique. Since all dot-products of~\eqref{eq:xent2} are unchanged by a joint rotation of $\nu(\x)$ and $\w_c$, the cross-entropy loss is invariant to rotations. In Figure~\ref{fig:intuition}, a rotation transforms the boundaries between classes from the coordinate axes on the left to the dashed lines on the right. From a classification perspective, the two solutions are identical.

For hashing, however, the two solutions are different. On the left, where each class occupies its own quadrant, all examples from the same similarity class share the same hash code $b(\x)$ (as defined in \eqref{eq:b}), while distinct hashes identify examples from different classes. This makes the hash codes optimal for retrieval. However, on the right, examples marked by a square have zero Hamming distance, despite belonging to different classes, and those marked by a star have distance one, despite belonging to the same class. In summary, while the two solutions are optimal for classification, only the one on the left is optimal for hashing. Thus, when the network of Figure~\ref{fig:deephash} is trained to optimize classification, the introduction of the $\tanh$ non-linearities in \eqref{eq:f} is not enough to guarantee good hash codes. Once the learning algorithm reaches the solution on the right of Figure~\ref{fig:intuition}, there is no classification benefit to pursuing that on the left. Since there is an infinite number of rotations that produce equally optimal solutions for classification, it is unlikely that the algorithm will ever produce the one optimal for hashing. 

\section{Learning Proxies for Hashing}
\label{sec:hashing-with-proxies}
In this section, we introduce a procedure to design proxies that induce good
hashing performance.

\subsection{Joint Optimality for Classification and Hashing}

So far we have seen that, because cross-entropy optimization leads to~(\ref{eq:hopt}), the set of proxies ultimately defines the properties of the learned embedding. This suggests that, rather than learning both embedding and proxies simultaneously, learning can proceed in two steps:

\begin{enumerate}
    \item Design a set of proxies $\w^*_y$ that encourages an embedding $\nu(\x)$ optimal for classification \textit{and} hashing;
    \item Learn the embedding $\nu^*(\x)$ by optimizing the CNN with the cross-entropy loss, while keeping the proxies $\w^*_y$ fixed.
\end{enumerate}
The ensuing question is ``which properties the set of proxies $\w_y$ must have to encourage optimal classification \textit{and} hashing?''

%%%%%%%%%%%%%%%%%%%%%%%%%%%%%%%%%%%%%%%%%%%%%%%%%%%%%%%%%%%%%%%%%

\paragraph{Proxies for optimal classification: }
To determine how the set of proxies can encourage optimal classification, we note that models learned by cross-entropy minimization are (approximately) max-margin classifiers. This can be seen by writing \eqref{eq:xent2} as
\begin{equation}
    \textstyle L_p(\x, y) = \log 
    \left[1+\sum_{c\neq y} e^{\langle \nu(\x), \w_c\rangle - \langle \nu(\x), \w_y\rangle}
    \right].
    \label{eq:xent3}
\end{equation}
Due to the exponent, the sum is dominated by the largest term, and minimizing \eqref{eq:xent3} is equivalent to minimizing $\max_{c\neq y} \langle \nu(\x), \w_c\rangle - \langle \nu(\x), \w_y\rangle$. Hence, the network seeks a predictor that maximizes the classification margin 
\begin{align}
    {\cal M}(\nu(\x), y) 
    & = \langle \nu(\x), \w_y\rangle - \max_{c \neq k} \langle \nu(\x), \w_c\rangle.
\end{align}
For the predictor of~(\ref{eq:hopt}), this is given by
\begin{align}
  {\cal M}(\nu^*(\x), y) 
  & \propto ||\w_y||^2 - \max_{c \neq k} \langle \w_y, \w_c\rangle.
\end{align}
Hence, to encourage classification optimality, it suffices to chose a set of fixed norm proxies, $||\w_c||^2 = K, c = \{1, \ldots, C\}$, that maximizes
\begin{equation}
    \textstyle {\cal M}_y = K - \max_{c \neq y} \left<\w_y,\w_c\right>,
    \label{eq:maxmargin}
\end{equation}
for all $y$, simultaneously. This is equivalent to solving 
\begin{equation}
    \begin{aligned}
        \WM^* =\ &
        \underset{\w_1,\ldots,\w_C}{\arg\max} & 
        \min_{i \neq j} ||\w_i - \w_j||^2 \\
        & \text{subject to} & ||\w_c||^2 = K \quad \forall c,
        \label{eq:tammes}
    \end{aligned}
\end{equation}
a classical problem in mathematics, known as the Tammes or sphere packing problem \cite{Tammes1930}, when $K=1$. The Tammes problem determines the maximum diameter of $C$ equal circles that can be placed on the surface of the unit sphere without overlap. In sum, a network trained to minimize classification loss encourages predictions aligned with proxies $\w_y$ for images of class $y$. Thus, classification margins are maximized when the proxy set is maximally separated, i.e., when the proxies are given by the Tammes solution $\WM^*$.

%%%%%%%%%%%%%%%%%%%%%%%%%%%%%%%%%%%%%%%%%%%%%%%%%%%%%%%%%%%%%%%%%

\paragraph{Hashing optimal proxies:}
Since embeddings $\nu^*(\x)$ cluster around the proxies of the corresponding
class $\w_y$, binary proxies encourage binary embedding representations.
Note that this is what is special about the solution of
Figure~\ref{fig:intuition} (left). Thus, joint optimality for classification
and hashing is guaranteed by any set of \textit{binary} proxies
that solve~\eqref{eq:tammes}, i.e.
\begin{equation}
	\begin{aligned}
    	\WM_h^* =\ &
    	\underset{\w_1,\ldots,\w_{C}}{\arg\max} & 
    	\min_{i \neq j} ||\w_i - \w_j||^2 \\
    	& \text{subject to} & \w_c \in \{-1,1\}^d \quad \forall c.
    	\label{eq:tammesbin}
	\end{aligned}
\end{equation}
Since learning the CNN with proxies $\WM_h^*$ as a
\textit{fixed} set of weights on the final softmax regression layer already
encourages a binary $\nu^*(\x)$, the CNN can be trained to
optimize classification only. The binarization step incurs no loss of
classification performance and there is no need to define additional cost
terms, which often conflict with classification optimality. Finally,
CNN optimization no longer has to deal with the ambiguity of multiple
solutions optimal for classification but not hashing.

%%%%%%%%%%%%%%%%%%%%%%%%%%%%%%%%%%%%%%%%%%%%%%%%%%%%%%%%%%%%%%%%%

\subsection{Proxy design}
It remains to determine a procedure to design the proxy set. This is not
trivial, since \eqref{eq:tammesbin} is a discrete optimization problem. In this work, we adopt an approximate solution composed of two steps. First, we solve the Tammes problem of~\eqref{eq:tammes} using a barrier method~\cite{Wright:Optimization} to obtain $\WM^*$, which we denote the Tammes proxies. Since the problem is convex, this optimization is guaranteed to produce a maximally separated proxy set. However, since any rotation around the origin leaves the norms of~\eqref{eq:tammes} unchanged, Tammes proxies are only defined up to a rotation. We exploit this degree of freedom to seek the rotation $\G^*$ that makes the Tammes proxies most binary. This consists of solving
\begin{eqnarray}
    \G^* &=& \arg \min_{\G \in SO_d} \textstyle\sum_k 
             \|\G \w_k^* - \sgn(\G \w_k^*)\|^2
    \label{eq:G*}
\end{eqnarray}
and is an instance of the binary quantization problem studied in~\cite{Gong:ITQ}. Given $\WM^*$, we use the ITQ binary quantization algorithm of~\cite{Gong:ITQ} to find the optimal rotation $\G^*$ of~\eqref{eq:G*}. It should be emphasized that, unlike previous uses of ITQ, the procedure is not used to binarize $\nu(\x)$, but to generate proxies that induce a binary $\nu(\x)$. This enables end-to-end training under a classification loss that seeks maximum discrimination between the classes used to define image similarity. Finally, the proxy matrix 
\begin{equation}
    \WM^*_h = \sgn(\G^*\WM^*).
    \label{eq:hashH*}
\end{equation}
is used to determine the weights of the softmax regression layer. We denote $\WM^*_h$ as the \textit{hash-consistent large margin} (HCLM) proxy set.

%%%%%%%%%%%%%%%%%%%%%%%%%%%%%%%%%%%%%%%%%%%%%%%%%%%%%%%%%%%%%%%%%

\subsection{Class/proxy matching}
\label{sec:semantic_proxies}
HCLM proxies $\w_y$ induce hash codes with good properties for both
classification and retrieval. 
However because $\w_y$ are simply a set of multi-dimensional class
labels, any permutation of the indices $y$ produces a set of valid
proxies.  This is probably best understood by referring to the four
class example of Figure~\ref{fig:intuition}. Assume that, in this example,
the classes were ``A: apples,'' ``C: cats,'' ``D: dogs,'' and
``O: oranges.'' While HCML generates the set of vectors $\w_y$, it does
not determine which vector in ${\cal W} = \{\w_1, \w_2, \w_3, \w_4\}$
should be paired with each label in ${\cal L} = \{A, C, D, O\}$.

While, in principle, with
enough data and computation, the embedding model $\nu(\cdot)$ could be
trained to map images $\x$ into the geometry induced by any pairing
between the elements of $\cal W$ and $\cal L$, some pairings
are easier to learn than others. Because learning is initialized with the
feature extractor $q(\x)$ from the  pre-trained network
(see Figure~\ref{fig:deephash}), the embedding $\nu(\cdot)$ should be
easier to learn when the pairing of proxies and classes respects the
similarity structure already available in this feature space. In the
example above, classes C and D will induce feature vectors in $q(\x)$
that are more similar than those of either the pair (C,A) or (D,A).
Similarly, the features vectors of classes A and O will be closer to each
other than to those of the other classes.

It follows that the pairing
\{($\w_1$, A), ($\w_2$,O), ($\w_3$, C), ($\w_4$, D)\},
where the vectors of similar classes are close to each other, respects
the structure of the feature space much better than the pairing
\{($\w_1$, A), ($\w_2$,C), ($\w_3$, O), ($\w_4,D)\}$, where they are opposite to
each other. More generaly, if classes $i$ and $j$ induce similar feature
vectors in $q(\x)$, which are both distant from the feature vectors induced
by class $z$, the assignment of proxies to classes should guarantee
that $d(\w_i,\w_j)$ is smaller  than $d(\w_i,\w_z)$ and $d(\w_j,\w_z)$.
This avoids the model $q(\x)$ to be required to {\it relearn} the structure
of the metric space, making training more data efficient, and
ultimately leading to a better hashing system.

Hence, the goal is to align the proxy similarities $\w_{y_i}^T\w_{y_j}$ with some measure of similarity $s_{ij}$ between feature vectors $q(\x)$
from classes $i$ and $j$.
This can be done by searching
for the proxy assignments $\gamma_1,\ldots,\gamma_C$ that minimize
\begin{equation}
    \textstyle\min_{\gamma_1,\ldots,\gamma_C}\quad
    \sum_{i\neq j}s_{ij}(1-\w_{\gamma_i}^T\w_{\gamma_j}^T).
    \label{eq:class_assignment}
\end{equation}
Although this is a combinatorial optimization problem, in our experience,
a simple greedy optimization is sufficient to produce a good solution.
Starting from a random class assignment, the proxy swap that leads to
the greatest decrease in~\eqref{eq:class_assignment} is taken at each
iteration, until no further improvement is achieved. This alignment
procedure is applied to HCLM to produce a {\it semantic HCLM\/} (sHCLM) proxy
set.
It remains to derive a procedure to measure the similarities $s_{ij}$ between classes.
We consider separately the cases where image similarity is derived from
single and multi labeled data.

{\it Single label similarity:}
For single labeled data, we first compute the average code ${\bf u}_y$ of class $y$, by averaging the feature vectors $q(\x_i)$ produced by the pre-trained network for training images $\x_i$ of class $y$
\begin{equation}
    {\bf u}_y = \frac{1}{n} \sum_{\x_i: y_i=y } q(\x_i).
    \label{eq:class_avg}
\end{equation}
Pairwise similarities between training classes $y_i$ and $y_j$ are 
then computed with
\begin{equation}
  \textstyle s_{ij} = 
  \exp\left\lbrace-\frac{\|{\bf u}_{y_i}-{\bf u}_{y_j}\|^2}{2\kappa^2}
  \right\rbrace, 
    \label{eq:similarity}
\end{equation}
where $\kappa$ is the average distance between means ${\bf u}_i$.
This procedure can also be justified by the fact that, as shown in
Appendix~\ref{app:equivalence} ~(\ref{eq:mu=w}) and
(\ref{eq:psi}), when $d(\cdot,\cdot)$ is the $L_2$ distance,
the proxy $\w_y$ is the average $\mu_y^\nu$ of the feature vectors
$\nu(\x_i)$ extracted from class $y$. However, because the class-proxy
assignments must be defined before training the network, the
embedding $\nu(\x)$ is not available. The use of (\ref{eq:class_avg})
corresponds to approximating the distance between vectors
$\nu(\x)$ by the distance between the vectors $q(\x)$ computable
with the pre-trained network. Note that slightly better performance could
likely be attained by first fine-tuning the pre-trained network on
the target dataset, without layers ${\bf L}$ and ${\bf W}^*$ of
Figure~\ref{fig:deephash}. However, this would increase training
complexity and is not used in this work.

{\it Multi label similarity:}
In multi labeled datasets, images are not described by a single class.
Instead, each image is annotated with $T$ auxiliary semantics (or tags) that
indicate the presence/absence of $T$ binary visual concepts. Each image $\x$ is
labeled with a binary vector $\bf t$, such that $t_k=1$ if the $k^{th}$ tag
is associated with the image and $t_k=0$ otherwise. 
In this case, proxies ${\bf w}_i$ are assigned to each tag $t_i$.

Given a multi-labeled dataset, (\ref{eq:class_avg}) and (\ref{eq:similarity}) 
could be used to compute tag similarities. 
However, we found experimentally that using tag co-occurrence as a 
measure of similarity can also yield strong performance.
Specifically, the similarity $s_{ij}$ between tags $t_i$ and $t_j$ was 
computed as
\begin{equation}
	s_{ij}=\frac{2\sum_n t_{n, i}\;t_{n, j}}{\sum_n t_{n, i}+\sum_n t_{n, j}},
\end{equation}
where $t_{n, i}\in\{0, 1\}$ denotes the $i^{th}$ tag of sample 
${\bf x}_n$. The similarity $s_{ij}$ approaches one when the 
$i^{th}$ and $j^{th}$ tags co-occur with high chance and is 
close to zero when the two tags never appear together.
This procedure has two advantages. 
First, it encourages tags that often co-occur to have similar 
proxies (i.e.~to share bits of the hash code).
Second, it has smaller complexity, since there is no need to 
forward images through the network to compute (\ref{eq:class_avg}).

%%%%%%%%%%%%%%%%%%%%%%%%%%%%%%%%%%%%%%%%%%%%%%%%%%%%%%%%%%%%%%%%

\subsection{Joint Proxy and Triplet Embedding}
\label{sec:joint_proxy_triplet}

So far, we have considered metric learning with proxy embeddings.
An alternative approach is to abandon the softmax
regression of~\eqref{eq:softmax} and apply a loss function directly
to $\nu(\x)$.  While many losses have been proposed~\cite{chopra2005,
  hadsell2006, sohn2016, oh2016}, the most popular operate on example
triplets, pulling together (pushing apart) similar (dissimilar)
examples~\cite{weinberger2009, wang2014, bell2015, schroff2015}.
These methods are commonly known as {\it triplet\/} embeddings.
When compared to proxy embeddings, they have both advantages and
shortcomings. On one hand, because the number of triplets
in the training set is usually very large, a subset of triplets
must be sampled for learning.
Despite the availability of many sampling
strategies \cite{schroff2015,wang2014,oh2016,sun2014}, it is usually
impossible to guarantee that the similarity information of the dataset is
fully captured. Furthermore, because they do not directly leverage
class labels, triplet embeddings tend to have weaker performance for
classification. On the other hand, because the similarity supervision
is spread throughout the feature space, rather than concentrated
along class proxies, they tend to better capture the metric structure of the
former away from the proxies. This more uniform learning of
metric structure is advantageous for applications such as 
retrieval or transfer learning, where triplet embeddings can outperform proxy
embeddings.

In this work, we found that combining proxy and triplet embeddings is often advantageous.
Given an anchor ${\bf x}$, a similar ${\bf x}^+$ and a dissimilar example ${\bf x}^-$, we define the triplet loss in Hamming space by the logistic loss with a margin of $m$ bits
\begin{equation}
    L_t(\x, \x^+, \x^-) = \log \left( 1+e^{m+d_H(\x,\x_p)-d_H(\x,\x_n)} \right).
    \label{eq:tl}
\end{equation}
Since, in hashing, $\nu(\x) \in [-1,1]$ represents a continuous surrogate of the hash codes $b(\x) \in \{0,1\}$, Hamming distances between two images $\x_i$ and $\x_j$ are estimated with the distance function
\begin{equation}
  d_H(\x_i, \x_j) = \frac{1}{2}\left(b-\nu(\x_i)^T\nu(\x_j)\right).
\end{equation}
Finally, given an sHCLM proxy set $\{\w_y\}_{y=1}^C$ of $C$ classes, the embedding is learned by 1) fixing the weights of the softmax regression layer to the proxies $\w_y$, and 2) learning the embedding $\nu(\x)$ to minimize
\begin{equation}
  L({\bf x}, y, {\bf x}^+, {\bf x}^-) =
  L_p({\bf x}, y) + \lambda L_t({\bf x}, {\bf x}^+, {\bf x}^-)
  \label{eq:jointloss}
\end{equation}
where $L_p$ is the proxy loss, $L_t$ given by~\eqref{eq:tl} 
and $\lambda$ is an hyper-parameter that controls their trade-off. 
In preliminary experiments, we found that the performance 
of joint embeddings is fairly insensitive to the value of $\lambda$. 
Unless otherwise noted, we use $\lambda = 1$ in all our experiments.
The proxy loss $L_p({\bf x}, y)$ is defined as the softmax cross-entropy 
loss \eqref{eq:xentrisk} in the case of single label similarities. 
For multi label similarities, to account for the fact that tag frequencies
are often imbalanced, we adopt a balanced binary cross-entropy~\cite{xie2015holistically}
\begin{eqnarray}
	\textstyle L_p({\bf x}, {\bf t}) &= - \textstyle \sum_k & c_k t_k \log s_k({\bf x}) + \\
	&&(1-c_k)(1-t_k) \log (1-s_k({\bf x})) \nonumber
\end{eqnarray}
where $s_k({\bf x})=\frac{1}{1+e^{{\bf w}_k^T \nu({\bf x})}}$ and $c_k$ is the inverse of the frequency of the $k^{th}$ tag .

\section{Experiments}
\label{sec:experiments}

In this section, we present an extensive experimental evaluation of the proposed hashing algorithm.

\subsection{Experimental setup}
A training set is used to learn the CNN embedding $\nu(\x)$, a set of images is defined as the image database and another set of images as the query database. Upon training, the goal is to rank database images by their similarity to each query. 

\paragraph{Datasets}
Experiments are performed on four datasets. 
CIFAR-10~\cite{CIFAR} and CIFAR-100~\cite{CIFAR} consist of 60\,000 color images (32$\times$32) from 10 and 100 image classes, respectively. 
Following the typical evaluation protocol~\cite{Wang:TripletSH, he2018hashing, Li:DeepSDH, Xia:CNNH}, we use the CIFAR test sets to create queries and the training sets for both training and retrieval databases. 
NUS-WIDE~\cite{NUS-WIDE} is a multi-label dataset composed of 270\,000 web images, annotated with multiple labels from a dictionary of 81 tags. 
Following standard practices for this dataset~\cite{Wang:TripletSH,Li:DPSH,Zhang:BitScalable,Li:DeepSDH}, we only consider images annotated with the 21 most frequent tags. 100 images are sampled per tag to construct the query set, and all remaining images are used both for training and as the retrieval database. 
ILSVRC-2012~\cite{Deng:Imagenet} is a subset of ImageNet with more than 1.2 million images of 1\,000 classes. In this case, the standard validation set (50\,000 images) is used to create queries, and the training set for learning and retrieval database. 

\paragraph{Image representation}
Unless otherwise specified, the base CNN of Figure~\ref{fig:deephash} is AlexNet~\cite{Krizhevsky:Alexnet} pre-trained on the ILSVRC 2012 training set and finetuned to the target dataset. Feature representations $q(\x)$ are the 4096-dimensional vectors extracted from the last fully connected layer before softmax regression (layer \textit{fc7}). On the CIFAR datasets, images are resized from the original 32$\times$32 into 227$\times$227 pixels, and random horizontal flipping is applied during training. On ILSVRC-2012 and NUS-Wide, images are first resized to 256$\times$256, and in addition to horizontal flipping, random crops are used for data augmentation. The central 227$\times$227 crop is used for testing.

\begin{figure*}[t!]%
    \centering
    \includegraphics[width=0.75\linewidth]{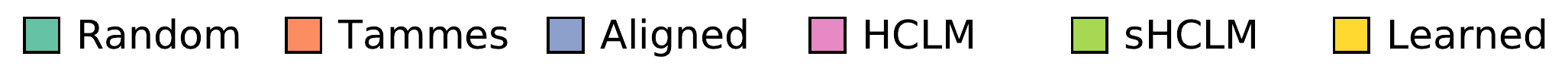}\\
    \subfloat[Classification margins. \label{fig:margin}]{
        {\includegraphics[width=0.49\linewidth]{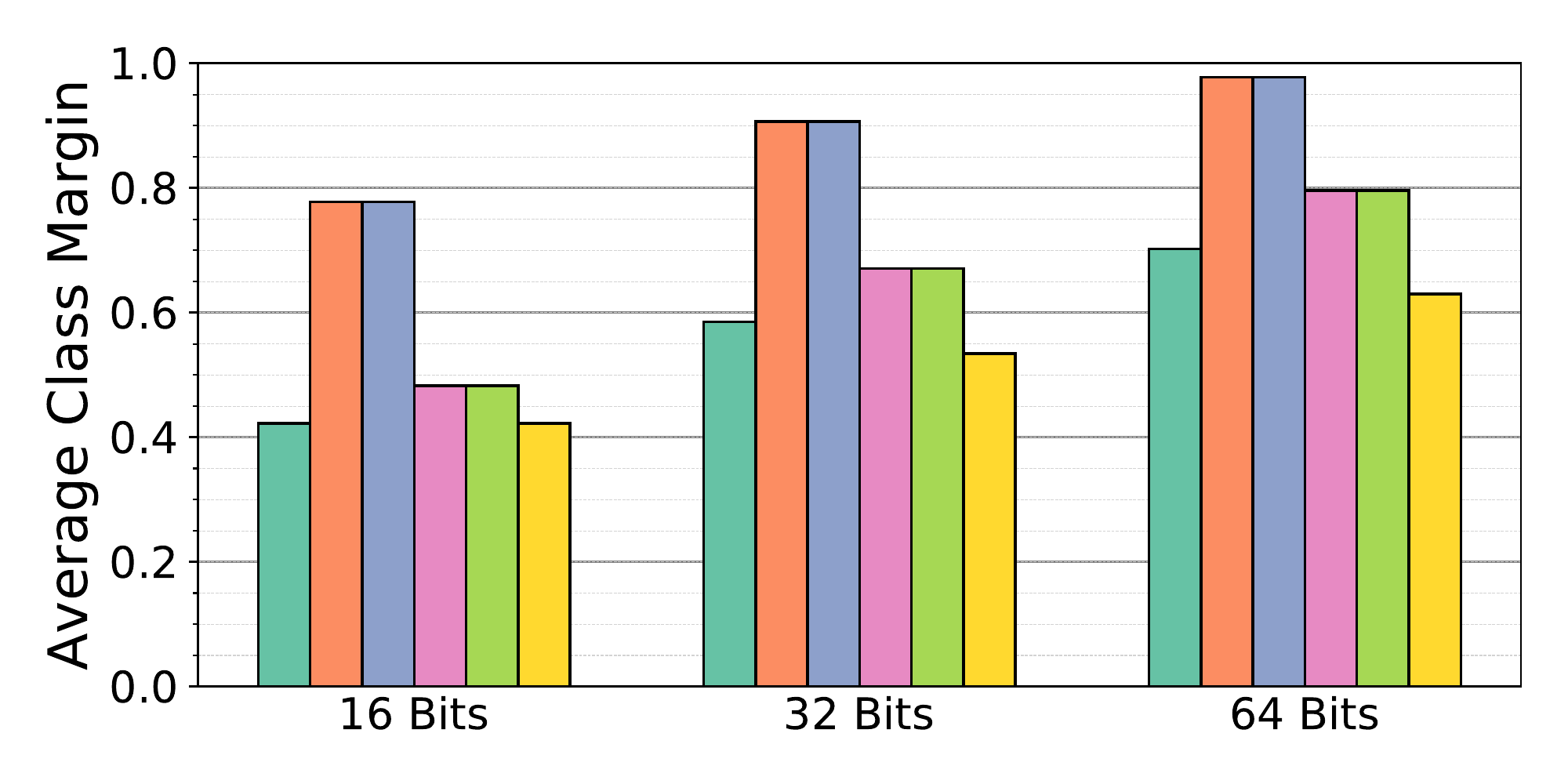}}%
    }
    \subfloat[Retrieval performance (mAP). \label{fig:versions}]{
        {\includegraphics[width=0.49\linewidth]{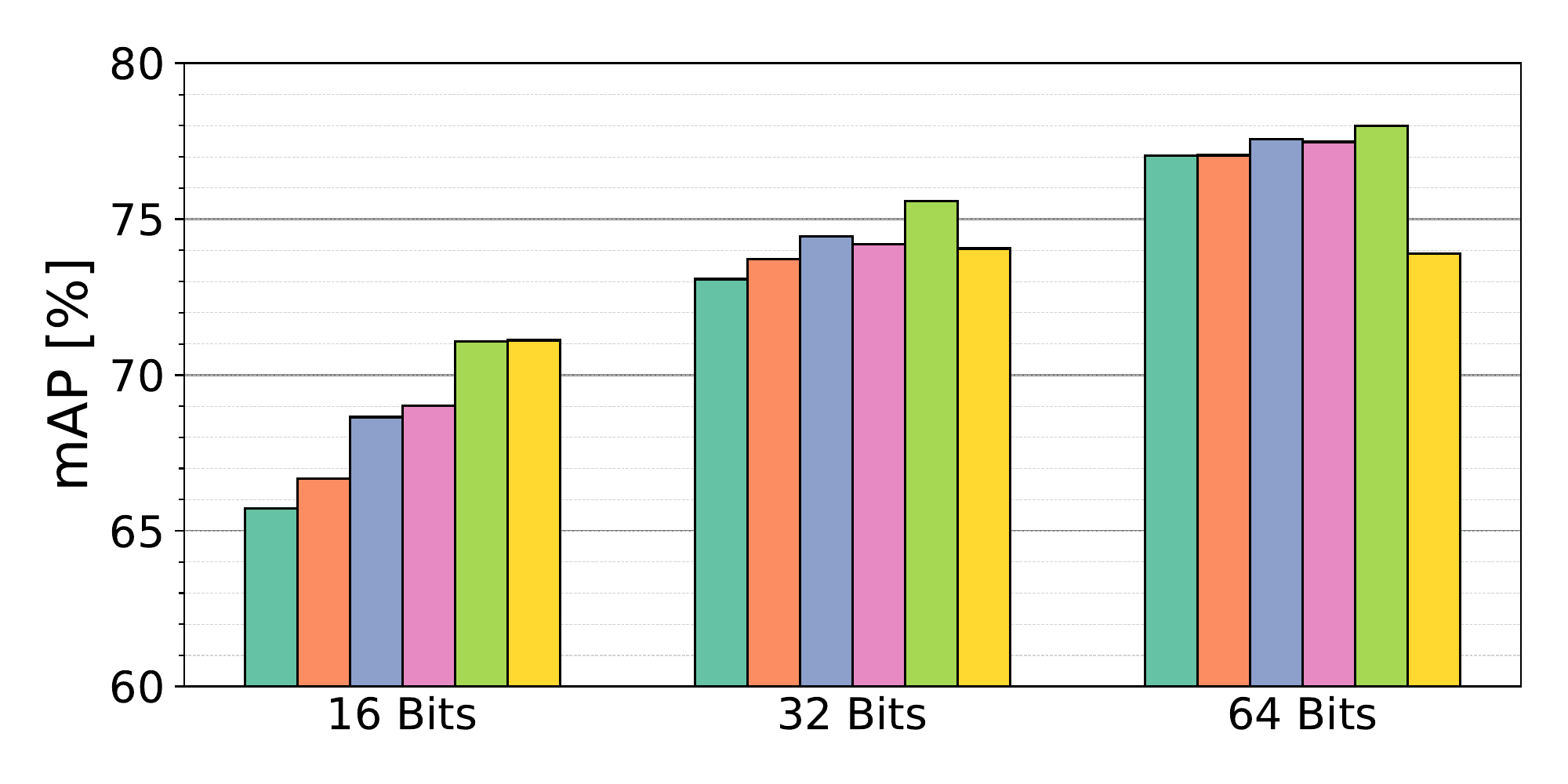}}
    }
	\caption{Classification and retrieval performance using different types of proxies on CIFAR-100.}
\end{figure*}

\begin{figure*}[t!]%
    \centering
    \subfloat[Histogram of Learned and sHCLM proxy weights. \label{fig:code_hist}]{
        {\includegraphics[height=1.8in]{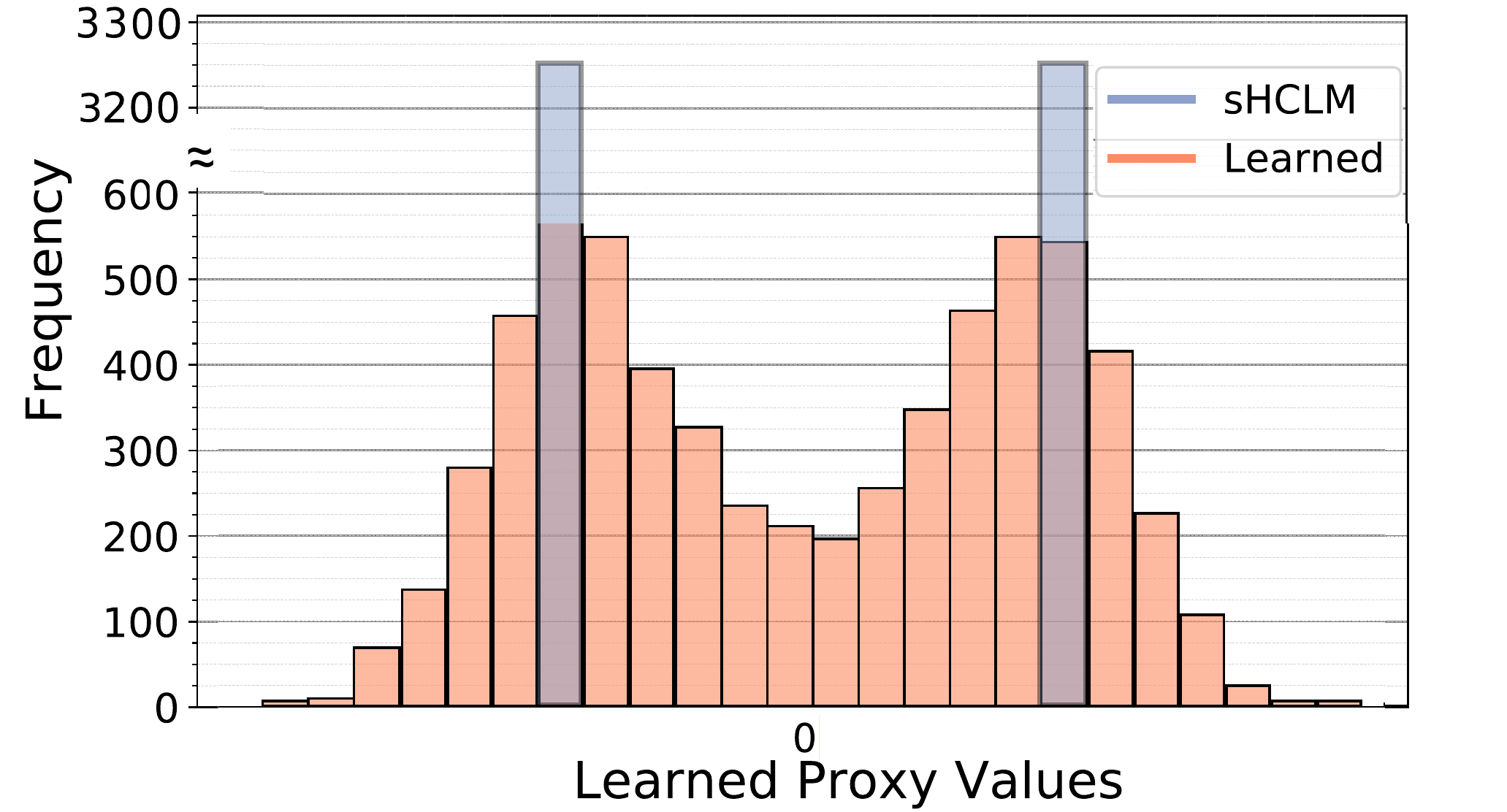}}
    }%
    \subfloat[Histogram of binarization errors. \label{fig:BinError}]{
        {\includegraphics[height=1.9in]{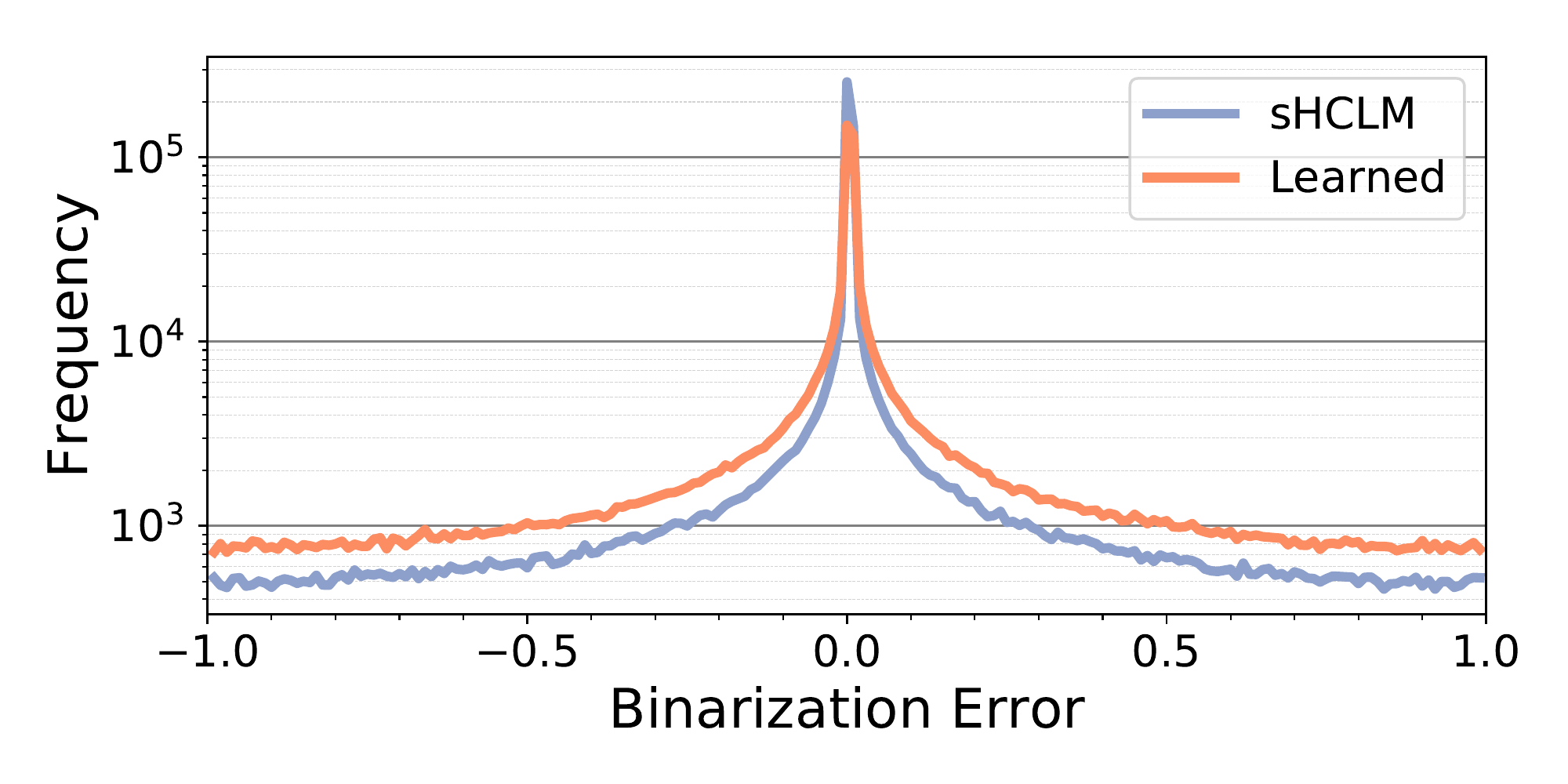}}
    }%
	\caption{Analysis of 64-bit proxy sets for hashing on CIFAR-100.}
	\label{fig:proxy_type}
\end{figure*}

\paragraph{Baselines}
Various methods from the literature were used for comparison, 1) classical 
(shallow) unsupervised algorithms, LSH~\cite{Datar:LSH} and ITQ~\cite{Gong:ITQ}; 
2) classical (shallow) supervised algorithms: SDH~\cite{Shen:SDH}, 
KSH~\cite{Liu:KSH} and ITQ with Canonical Correlation Analysis 
(ITQ-CCA)~\cite{Gong:ITQ}; and 3) deep supervised hashing algorithms: 
DQN~\cite{Cao:DQN}, CNNH~\cite{Xia:CNNH}, NINH~\cite{Lai:NINH}, DSRH~\cite{Zhao:DSRH}, 
DRSCH~\cite{Zhang:BitScalable}, SSDH~\cite{Yang:SSDH}, DPSH~\cite{Li:DPSH}, 
SUBIC~\cite{jain2017subic}, BHC~\cite{Lin:BHC}, DTSH~\cite{Wang:TripletSH}, 
MI-Hash~\cite{Cakir:MIHash}, TALR-AP~\cite{he2018hashing}, DSDH~\cite{Li:DeepSDH}, 
HBMP~\cite{cakir2018hashing} either based on weighted Hamming distances (denoted 
``HBMP regress'' in~\cite{cakir2018hashing}) or binary Hamming distances (denoted 
``HBMP constant'' in~\cite{cakir2018hashing}), and ADSH~\cite{jiang2018asymmetric}.
We restate published results when available. Author implementations with default 
parameters were used for LSH, ITQ, ITQ-CCA, SDH, and KSH with off-the-shelf 
AlexNet features. We also report results for BHC, SSDH and DTSH for experimental settings not 
considered in the original paper. To ensure representative performances, we first 
reproduced published results, typically on CIFAR-10, using author implementations, 
and then apply the same procedure to the target dataset.

\paragraph{Evaluation metrics}
Retrieval performance is evaluated using the mean of the average precision
(mAP) across queries. Given a ranked list of $n$ database matches to a query
image $\x$, the aggregate precision of the top-$k$ results, over
all cutoffs $k$, is computed with
\begin{equation}
AP=\sum_{k=1}^n P(k)\Delta r(k)
\end{equation}
where $P(k)$ is the precision at cutoff $k$, and $\Delta r(k)$ the change in
recall from matches $k-1$ to $k$.
For both NUS-WIDE and ImageNet, only the top $n=50\,000$ retrievals are
considered when computing the AP. For CIFAR-10 and CIFAR-100, AP is
computed over the full ranking. The mAP is the average AP value over the
set of queries.

%%%%%%%%%%%%%%%%%%%%%%%%%%%%%%%%%%%%%%%%%%%%%%%%%%%%%%%%%%%%%%%%%%%

\subsection{Learning to hash with explicit class similarities}
\label{sec:supervised_results}
We start by evaluating hashing performance on datasets where image similarity is directly derived from class labels (CIFAR-10, CIFAR-100, and ISLVRC-2012). 

\subsubsection{Ablation study}
\label{sec:analysis}
We start by ablating the proxy generation procedure of Figure~\ref{fig:deephash}, using the CIFAR-100 dataset. Several embeddings were trained on this dataset, each using a different proxy version. The proxies are as follows.
\begin{enumerate}
\item {\bf Learned:} proxy set learned by back-propagation; 
\item {\bf Random:} fixed random proxies; 
\item {\bf Tammes:} optimal solution $\WM^*$ to the Tammes problem of~\eqref{eq:tammes}; 
\item {\bf Aligned:} proxy set $\G^*\WM^*$ obtained by rotating $\WM^*$ with the rotation matrix $\G^*$ of~\eqref{eq:G*};
\item {\bf HCLM:} HCLM proxy set $\WM_h^*$ of~\eqref{eq:hashH*} (with random proxy/class assignments);
\item {\bf sHCLM:} HCLM set after the semantic proxy/class assignment of~\eqref{eq:class_assignment}.
\end{enumerate}

Figure~\ref{fig:margin} shows the margin associated with each proxy set (computed with~\eqref{eq:maxmargin}) averaged across classes. Figure~\ref{fig:versions} compares the retrieval performance of the resulting hash codes. These results support several conclusions. First, as expected, Tammes produces the largest classification margins. However, because it disregards the binarization requirements of hashing, it induces an embedding of relatively weak retrieval performance. 
This can also be seen by the retrieval gains of Aligned over Tammes. Since the two proxy sets differ only by a rotation, they have equal classification margins. However, due to \eqref{eq:hopt}, ``more binary'' proxies force more saturated responses of the $\tanh(\cdot)$ non-linearities and smaller binarization error. This improves the retrieval performance of the embedding. Second, while the binarization step of~\eqref{eq:hashH*} reduces classification margins, it does not affect retrieval performance. In fact, the HCLM embedding has better retrieval performance than the Aligned embedding for hash codes of small length. Third, the mAP gains of sHCLM over HCLM show that explicitly inducing a semantic embedding, which maps semantically related images to similar hash codes, further improves retrieval performance. Overall, the sHCLM embedding has the best retrieval performance. Finally, the Learned embedding significantly underperforms the sHCLM embedding. This shows that, at least for hashing, proxy embeddings cannot be effectively learned by back-propagation. In fact, for large code lengths, Learned proxies underperformed Random proxies in terms of both class margins and retrieval performance. As discussed in Sec.~\ref{sec:challenges}, this is explained by the fact that the cross-entropy loss is invariant to proxy rotations and most rotations do not induce low binarization error. This compromises the effectiveness of proxies learned by back-propagation for the hashing scenario.

To better quantify this issue, we compared the sHCLM and Learned proxy sets in more detail for hash codes of 64 bits. Figure~\ref{fig:code_hist} shows the histogram of weight values for the learned and sHCLM proxies. It is clear that, even when the proxy set is learned, the weights are bimodal. This is due to the inclusion of the $\tanh(\cdot)$ non-linearities at the output of $\nu(\x)$ and \eqref{eq:hopt}. However, unlike the sHCLM proxy sets (which are binary by construction), the weight distribution exhibits significant dispersion around the two modes. Since learned proxies are less binary than sHCLM, the same holds for the embedding $\nu(\x)$. This is confirmed by the binarization error histograms of Figure~\ref{fig:BinError}, which show a more binary embedding for sHCLM. In result, even though the sHCLM and Learned embeddings have nearly equal classification accuracy on CIFAR-100 (75.5\% and 75.4\%, respectively), sHCLM substantially outperforms the Learned proxy set for retrieval (Figure~\ref{fig:versions}).

\begin{table}[t!]
\centering
\caption{Retrieval performance (mAP) under supervised protocol on CIFAR-10. ${}^*$ Self-implementation}
\label{tab:supervised-CIFAR10}
\renewcommand{\arraystretch}{1.15}
\begin{tabular}{|r|cccc|}
    \cline{2-5}
    \multicolumn{1}{c|}{} & \multicolumn{4}{c|}{Hash size} \\
    \multicolumn{1}{c|}{} & 16 Bits & 24 Bits & 32 Bits & 48 Bits \\
    \hline
    \multicolumn{5}{|c|}{Classical} \\
    \hline
    LSH~\cite{Datar:LSH} & 17.5 & 20.2 & 20.4 & 21.2 \\
    ITQ~\cite{Gong:ITQ}  & 22.9 & 24.3 & 24.8 & 25.6 \\
    KSH~\cite{Liu:KSH}   & 47.8 & 50.5 & 50.4 & 52.8 \\
    SDH~\cite{Shen:SDH}  & 66.5 & 68.3 & 70.0 & 71.3 \\
    ITQ-CCA~\cite{Gong:ITQ}& 71.4 & 72.6 & 74.1 & 74.8 \\
    \hline
    \multicolumn{5}{|c|}{Proxy embeddings} \\
    \hline
    SUBIC~\cite{jain2017subic} & 63.5 & 67.2 & 68.2 & 68.6 \\
    %BHC~\cite{Lin:BHC} & 90.6 & --- & 91.2 & 91.5 \\
    BHC*~\cite{Lin:BHC} & 93.3 & 93.6 & 94.0 & 94.0 \\
    %SSDH~\cite{Yang:SSDH} & 90.6 & --- & 91.2 & 91.5 \\
    SSDH*~\cite{Yang:SSDH} & 93.6 & 93.9 & 94.2 & 94.1 \\
    \hline
    \multicolumn{5}{|c|}{Pair-wise embeddings} \\
    \hline
    DQN~\cite{Cao:DQN}	& --- & 55.8 & 56.4 & 58.0  \\
    DPSH~\cite{Li:DPSH}& 76.3 & 78.1 & 79.5 & 80.7 \\
    HBMP~\cite{cakir2018hashing} & 94.2 & 94.4 & 94.5 & 94.6 \\
    ADSH~\cite{jiang2018asymmetric} & 89.0 & 92.8 & 93.1 & 93.9 \\
    \hline
    \multicolumn{5}{|c|}{Triplet embeddings} \\
    \hline
    NINH~\cite{Lai:NINH}& --- & 56.6 & 55.8 & 58.1 \\
    DRSCH~\cite{Zhang:BitScalable}& 61.5 & 62.2 & 62.9 & 63.1 \\
    DTSH~\cite{Wang:TripletSH}	& 91.5 & 92.3 & 92.5 & 92.6 \\
    \hline
    \multicolumn{5}{|c|}{Ranking embeddings} \\
    \hline
    DSRH~\cite{Zhao:DSRH}& 60.8 & 61.1 & 61.7 & 61.8 \\
    MI-Hash~\cite{Cakir:MIHash}& 92.9 & 93.3 & 93.8 & 94.2 \\
    TALR-AP~\cite{he2018hashing}& 93.9 & 94.1 & 94.3 & 94.5 \\
    \hline
    \multicolumn{5}{|c|}{Combinations} \\
    \hline
    CNNH~\cite{Xia:CNNH} & 55.2 & 56.6 & 55.8 & 58.1 \\
    DSDH~\cite{Li:DeepSDH} & 93.5 & 94.0 & 93.9 & 93.9 \\
    \hline
    \multicolumn{5}{|c|}{Proposed} \\
    \hline
    sHCLM& \bf 94.5 & \bf 94.7 & \bf 95.2 & 94.9 \\
    sHCLM + Triplet & \bf 94.5 & \bf 94.7 & 94.9 & \bf 95.0 \\
    \hline
\end{tabular}
\end{table}

\subsubsection{Comparison to previous work}
We compared the hashing performance of various embeddings on CIFAR-10, 
CIFAR-100 and ImageNet.

\paragraph{\it CIFAR-10:} Since CIFAR-10 is one of the most popular 
benchmarks for hashing, it enables a more extensive comparison. 
Table~\ref{tab:supervised-CIFAR10} restates the performance of various 
methods as reported in the original papers, when available. 
The exceptions are classical hashing algorithms, namely 
LSH~\cite{Datar:LSH}, ITQ~\cite{Gong:ITQ}, SDH~\cite{Shen:SDH}, 
KSH~\cite{Liu:KSH} and ITQ with Canonical Correlation Analysis 
(ITQ-CCA)~\cite{Gong:ITQ}. In these cases, we used author 
implementations with default parameters and off-the-shelf AlexNet fc7 
features as input image representations. We also compare to 
several representatives of the deep hashing literature, including triplet 
embeddings (NINH~\cite{Lai:NINH}, DRSCH~\cite{Zhang:BitScalable} and 
DTSH~\cite{Wang:TripletSH}), pairwise embeddings (DQN~\cite{Cao:DQN}, 
DPSH~\cite{Li:DPSH}, HBMP~\cite{cakir2018hashing},
and ADSH~\cite{jiang2018asymmetric}), methods that 
optimize ranking metrics (DSRH~\cite{Zhao:DSRH}, TALR-AP~\cite{he2018hashing} 
and MI-Hash~\cite{Cakir:MIHash}), proxy embedding methods 
(SUBIC~\cite{jain2017subic}, SSDH~\cite{Yang:SSDH}, and
BHC~\cite{Lin:BHC}), or different combinations of these 
categories (CNNH~\cite{Xia:CNNH} and DSDH~\cite{Li:DeepSDH}).

Table~\ref{tab:supervised-CIFAR10} supports several conclusions. 
First, sHCLM achieves state-of-the-art performance on this dataset. 
sHCLM outperforms previous proxy embeddings. 
This is mostly because these methods complement the cross-entropy 
loss with regularization terms meant to reduce binarization error. 
However, it is difficult to achieve a good trade-off between the two 
goals with loss-based regularization alone. In contrast, because the 
sHCLM proxy set is nearly optimal for both classification and hashing, 
the sHCLM embedding can be learned with no additional binarization 
loss terms. This enables a much better embedding for hashing.
Second, most triplet and pairwise embeddings are also much less effective
than sHCLM, with only HBMP and ADSH achieving comparable performance. 
It should be noted, however, that both these approaches leverage non-binary
operations for retrieval (\cite{cakir2018hashing} uses weighted
Hamming distances, and \cite{jiang2018asymmetric} only binarizes database
images and uses full-precision codes for the queries). Hence, the
comparison to the strictly binary sHCLM is not fair.
Since sHCLM is compatible with any metric, it would likely also benefit
from the floating-point retrieval strategies of \cite{cakir2018hashing}
and \cite{jiang2018asymmetric}. Nevertheless, sHCLM
still outperforms the best of these methods (HBMP) by 0.6\%.
Third, among strictly binary methods, only the ranking and combined
embeddings achieve performance comparable, although inferior to sHCLM and,
as frequently observed in the literature, classical methods
cannot compete with deep learning approaches. 
Finally, the combination of the proxy sHCLM and triplet embedding 
has no noticeable performance increase over sHCLM alone. 
This suggests that, on CIFAR-10, there is no benefit in using 
anything more sophisticated than the sHCLM proxy embedding trained with 
cross-entropy loss.

\paragraph{\it CIFAR-100 and ILSVRC-2012:} While widely used, CIFAR-10 is
a relatively easy dataset, since it does not 
require dimensionality reduction, one of the main challenges of 
hashing. Because the hash code length $d$ is much larger than the number 
of classes ($C=10$), any good classifier can be adapted to hashing without 
significant performance loss. 
Learning hashing functions is much harder when $C>>d$, as is the case for 
CIFAR-100 and ILSVRC-2012.
In these cases, beyond classical methods, we compared to
the methods that produced the best results on CIFAR-10
(Table~\ref{tab:supervised-CIFAR10}) among proxy (BHC and SSDH) and triplet (DTSH)
embeddings. BHC~\cite{Lin:BHC} learns a CNN classifier with a sigmoid
activated hashing layer, SSDH~\cite{Yang:SSDH} imposes additional binarization 
constrains over BHC, and DTSH~\cite{Wang:TripletSH} adopts a binary triplet 
embedding approach.
Since they were not evaluated on CIFAR-100 and ILSRVC-2012, we used 
the code released by the authors.

Tables~\ref{tab:supervised-CIFAR100} and \ref{tab:supervised-ImageNet} 
show that the gains of sHCLM are much larger in this case, outperforming 
all methods by $1.7\%$ mAP points on CIFAR-100 (64 bits) and $3.5\%$ on 
ILSVRC 2012 (128 bits), with larger margins for smaller code sizes. 
The sampling difficulties of triplet-based approaches like
DTSH are evident for these datasets. 
Since the complexity of the similarity structure increases with the 
number of classes, sampling informative triplets becomes increasingly 
harder. In result, triplet embeddings can have very weak performance. 
The gains of sHCLM over previous proxy embeddings (BHC and SSDH) are also 
larger on CIFAR-100 and ILSVRC 2012 than CIFAR-10. Proxy embeddings are
harder to learn when $C$ is large because 
the network has to pack more discriminant power in the same $d$ bits of 
the hash code. When $C>>d$, it is impossible for the hashing layer 
$\nu(\cdot)$ to retain all semantic information in $f(\cdot)$. Hence, 
the CNN must perform \textit{discriminant dimensionality reduction}. 
By providing a proxy set already optimal for classification and hashing 
in the $d$-dimensional space, sHCLM faces a much simpler optimization 
problem. This translates into more reliable retrieval. Finally, the
gains of adding the triplet loss to sHCML were small or non-existent on these
datasets.

\begin{table}[t!]
\centering
\caption{Retrieval performance (mAP) under supervised protocol on CIFAR-100.}
\label{tab:supervised-CIFAR100}
\renewcommand{\arraystretch}{1.15}
\begin{tabular}{|r|ccc|}
    \cline{2-4}
    \multicolumn{1}{c|}{} & \multicolumn{3}{c|}{Hash size} \\
    \multicolumn{1}{c|}{} & 16 Bits & 32 Bits & 64 Bits \\
    \hline
    \multicolumn{4}{|c|}{Classical} \\
    \hline
    LSH~\cite{Datar:LSH}& 3.4 & 4.7 & 6.4 \\
    ITQ~\cite{Gong:ITQ} & 5.2 & 7.1 & 9.0 \\
    KSH~\cite{Liu:KSH}	& 9.3 & 12.9 & 15.8 \\
    SDH~\cite{Shen:SDH}& 19.1 & 25.4 & 31.1 \\
    ITQ-CCA~\cite{Gong:ITQ}& 14.2 & 25.0 & 33.8 \\
    \hline
    \multicolumn{4}{|c|}{Proxy embeddings} \\
    \hline
    BHC~\cite{Lin:BHC}    			& 64.4 & 73.7 & 76.2 \\
    % BHC (tanh)~\cite{Lin:BHC}	    & 70.1 & 74.1 & 73.9 \\
    SSDH~\cite{Yang:SSDH} 			& 64.6 & 73.6 & 76.6 \\
    %SSDH (tanh)~\cite{Yang:SSDH}    & 70.6 & 73.9 & 74.2 \\
    \hline
    \multicolumn{4}{|c|}{Triplet Embeddings} \\
    \hline
    DTSH~\cite{Wang:TripletSH}	& 27.6 & 41.3 & 47.6 \\
    \hline
    \multicolumn{4}{|c|}{Proposed} \\
    \hline
    sHCLM           &     71.1 &     75.6 &     78.0 \\
    sHCLM + Triplet & \bf 71.6 & \bf 76.1 & \bf 78.3 \\
    \hline
\end{tabular}
\end{table}

\begin{table}[t!]
\centering
\caption{Retrieval performance (mAP) under supervised protocol on ILSVRC-2012.}
\label{tab:supervised-ImageNet}
\renewcommand{\arraystretch}{1.15}
\begin{tabular}{|r|ccc|}
    \cline{2-4}
    \multicolumn{1}{c|}{} & \multicolumn{3}{c|}{Hash size} \\
    \multicolumn{1}{c|}{} & 32 Bits & 64 Bits & 128 Bits \\
    \hline
    \multicolumn{4}{|c|}{Classical} \\
    \hline
    LSH~\cite{Datar:LSH}	& 2.7 & 4.8 & 7.2 \\
    ITQ~\cite{Gong:ITQ} 	& 4.2 & 6.5 & 8.3 \\
    ITQ-CCA~\cite{Gong:ITQ}	& 5.0 & 9.1 & 13.8 \\
    \hline
    \multicolumn{4}{|c|}{Proxy embeddings} \\
    \hline
    BHC~\cite{Lin:BHC}	& 14.4 & 21.1 & 25.4 \\
    SSDH~\cite{Yang:SSDH} 			& 14.5 & 23.6 & 28.9 \\
    \hline
    \multicolumn{4}{|c|}{Triplet embedding} \\
    \hline
    DTSH~\cite{Wang:TripletSH}	& 6.1 & 8.0 & 12.2 \\
    \hline
    \multicolumn{4}{|c|}{Proposed} \\
    \hline
    sHCLM           & \bf 24.7 & \bf 29.9 & \bf 32.4 \\
    sHCLM + Triplet &     23.0 &     29.0 &     32.1 \\
    \hline
\end{tabular}
\end{table}

%%%%%%%%%%%%%%%%%%%%%%%%%%%%%%%%%%%%%%%%%%%%%%%%%%%%%%%%%%%%%%%%

\paragraph{Conclusions:} The experiments of this section show that, when
similarity ground truth is derived from the class labels
used for network training, sHCML outperformed all previous approaches
in the literature. In this setting, triplet and pairwise embeddings have
much weaker performance than proxy embeddings. Even the combination of
the two approaches, by addition of a triplet loss to sHCML, has minimal
improvements over sHCML. Compared to the previous proxy-based methods,
sHCML shows substantial gains for hashing on the more challenging datasets, where
the number of classes is much larger than the number of dimensions.

\begin{figure*}[t!]%
    \centering
    \subfloat[Retrieval performance (mAP). \label{fig:versions-transfer}]{
        \includegraphics[height=1.8in]{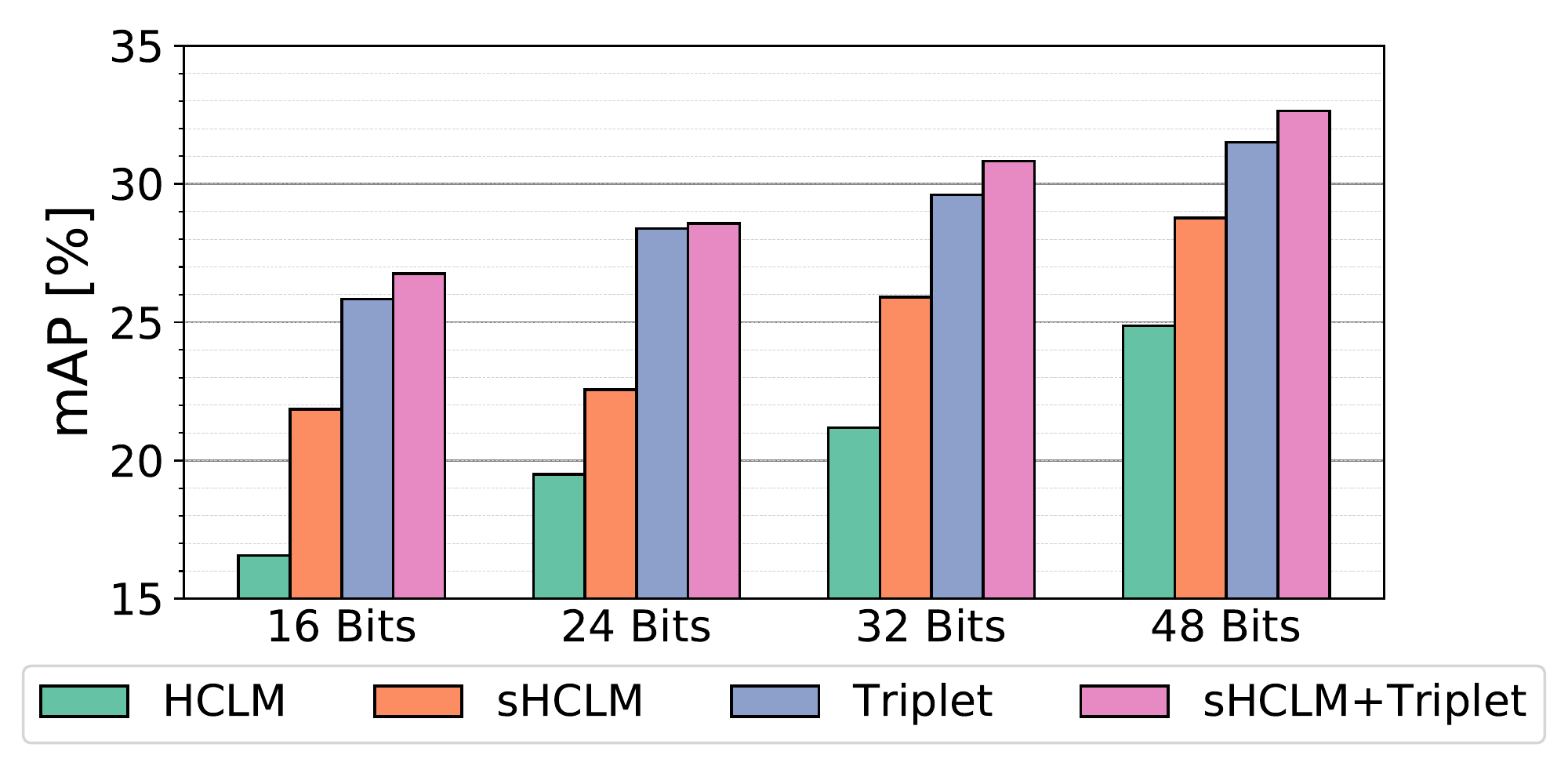}
    } \quad
    \subfloat[Precision at K. \label{fig:transfer-prec@k}]{
        {\includegraphics[height=1.8in]{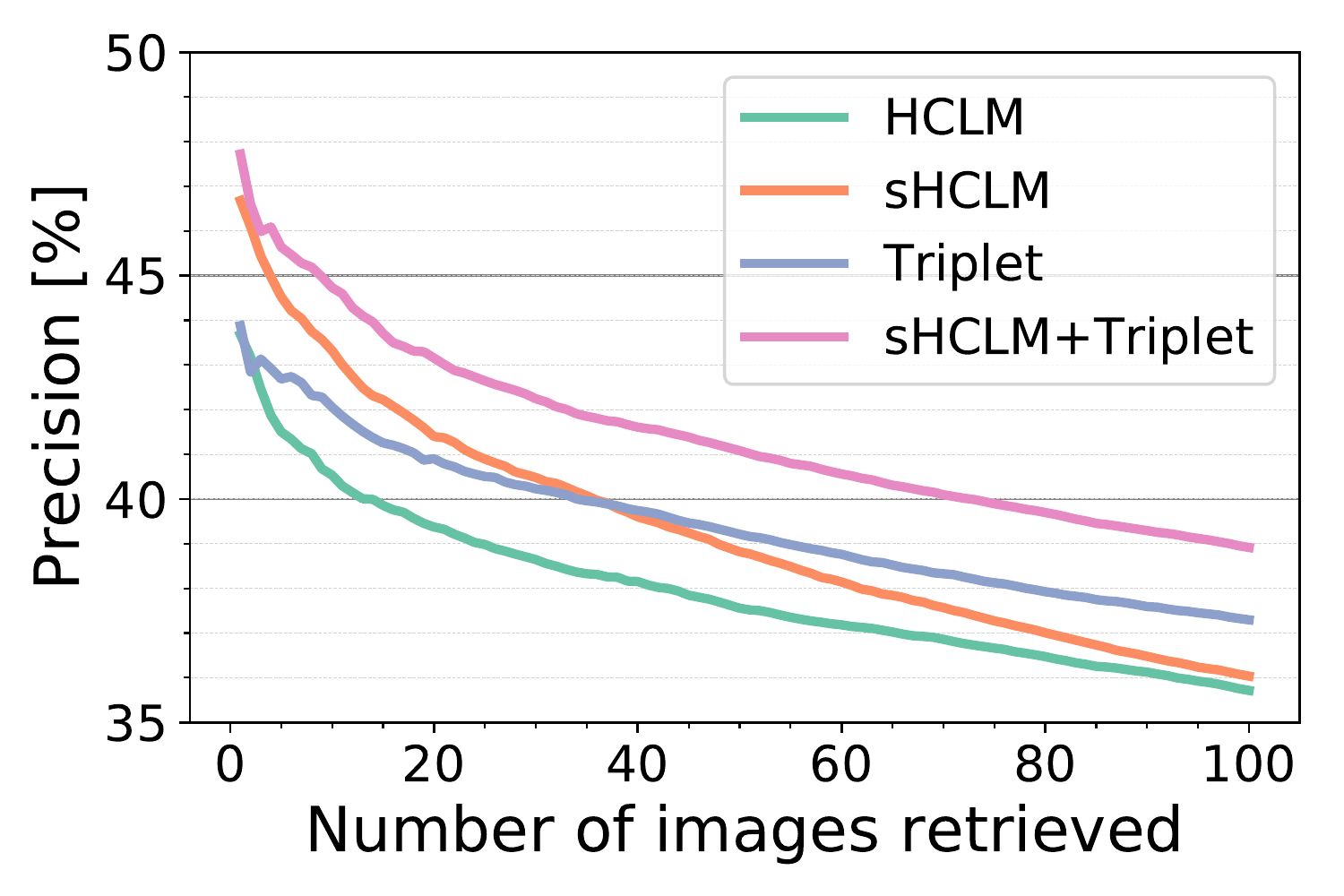}}
    }
	\caption{Retrieval performance on 25 unseen classes of the CIFAR-100 dataset using different versions of the proposed method.}
\end{figure*}

\subsection{Transfer Learning Performance}
We next evaluated performance under the transfer setting
of~\cite{Sablayrolles:SHEval}. 
This consists of learning the embedding with one set of classes and 
evaluating its retrieval and classification performance on a disjoint set.
Following~\cite{Sablayrolles:SHEval}, a 75\%/25\% class split was 
first defined. A set of training images was extracted from the 75\% split 
and used to learn the embedding. The sHCLM proxies of 
Section~\ref{sec:semantic_proxies} are only defined for these training classes, 
for which images were available to compute the class representations 
${\bf u}_y$ of~(\ref{eq:class_avg}). After training, hash codes $b(\x)$ were 
computed for the images of the 25\% split, by forwarding the images through 
the network and thresholding $\nu(\x)$. The quality of these hash codes 
was then evaluated in both retrieval and classification settings. For this, 
the hash codes were first divided into a database and a set of queries. 
Then, in the retrieval experiment, the database codes were ranked by 
similarity to each query. In the classification experiment, the database 
hash codes were used to train a softmax regression classifier, whose 
classification performance was evaluated on the query hash codes. 
As suggested in~\cite{Sablayrolles:SHEval}, all experiments were repeated 
over multiple 75\%/25\% splits. This was done by splitting the entire 
dataset into four 25\% sets of disjoint classes and grouping them into 
four 75\%/25\% splits, in a leave-one-out fashion. 
All results were averaged over these four splits.

%%%%%%%%%%%%%%%%%%%%%%%%%%%%%%%%%%%%%%%%%%%%%%%%%%%%%%%%%%%%%%%%

\subsubsection{Ablation Study}
Figures~\ref{fig:versions-transfer} and \ref{fig:transfer-prec@k} compare 
the retrieval performance of four embeddings. ``HCLM'' and ``sHCLM'' are 
proxy embeddings trained with the HCLM and sHCLM proxies, respectively. 
``Triplet'' is a triplet embedding, learned with the loss of~\eqref{eq:tl}, 
and ``sHCLM+Triplet'' is trained with the joint loss of~\eqref{eq:jointloss}, 
using sHCLM proxies. Figure~\ref{fig:versions-transfer} shows 
the mAP of the Hamming rankings produced by the different embeddings 
on the CIFAR-100 dataset. A comparison to Figure~\ref{fig:versions} shows 
that the gains of semantic class alignment (sHCLM) over the HCLM proxy set 
increase when the retrieval system has to generalize to unseen classes. 
This is in line with observations from the zero- and few-shot learning 
literature, where it is known that capturing semantic relationships between 
classes is critical for 
generalization~\cite{Lampert:Attributes, Akata:ALE, Morgado:SCoRe}.

However, the proxy embeddings tend to overfit on the training classes, 
under-performing the triplet embedding. This confirms previous findings
from the embedding literature, where triplet embeddings are known to
generalize better for applications that rely heavily on transfer, such as face
identification~\cite{schroff2015,wang2014}.
Note that the addition of  the triplet loss, which did not significantly
improve retrieval in experiments with the same training and test classes
(Tables~\ref{tab:supervised-CIFAR100} and~\ref{tab:supervised-ImageNet}), 
has a significant impact in the transfer setting of
Figure~\ref{fig:versions-transfer}. The increased
robustness of the triplet embedding against overfitting to the training 
classes is also evident in Figure~\ref{fig:transfer-prec@k}, which shows 
the average precision of the top $K$ retrieved images, for $32$-bits hash 
codes, as a function of $K$. Note that the sHCLM proxy embedding produces 
higher quality rankings for low values of $K$ than the triplet embedding. 
On the other hand, the latter has higher precision for large values of $K$. 
This suggests that, while the proxy embedding produces a better local 
clustering of the image classes, it underperforms the triplet embedding 
in the grouping of less similar images from each class. Overall, the 
combination of the sHCLM proxy and triplet embeddings achieves the best 
performance. This shows that the two approaches are complementary and 
there are benefits to an embedding based on their combination, even in 
the transfer setting.

\begin{table}[t!]
    \centering
    \caption{Retrieval performance (mAP)  under transfer protocol on CIFAR-100 and ILSVRC-2012 datasets.}
    \label{tab:transfer-retr}
    \renewcommand{\arraystretch}{1.15}
	\begin{tabular}{|r|ccc|}
		\hline\rule{0pt}{1.1\normalbaselineskip}
		Dataset & \multicolumn{3}{c|}{CIFAR-100} \\
		\# Bits & 16 bits & 32 bits & 64 bits \\
		\hline
	    \multicolumn{4}{|c|}{Classical} \\
		\hline
		LSH~\cite{Datar:LSH} & 11.0 & 14.2 & 16.6 \\
		ITQ~\cite{Gong:ITQ} & 14.4 & 17.3 & 20.1 \\
		KSH~\cite{Liu:KSH} & 16.2 & 17.1 & 18.4 \\
		SDH~\cite{Shen:SDH} & 12.6 & 15.5 & 14.8 \\
		ITQ-CCA~\cite{Gong:ITQ} & 17.5 & 18.1 & 18.2 \\
		\hline
	    \multicolumn{4}{|c|}{Proxy embeddings} \\
		\hline
		Softmax-PQ~\cite{Sablayrolles:SHEval} & --- & 22.0 & --- \\
		BHC~\cite{Lin:BHC} & 21.9 & 27.9 & 31.7 \\
		\hline
	    \multicolumn{4}{|c|}{Triplet embeddings} \\
		\hline
		DTSH~\cite{Wang:TripletSH} & 25.5 & 29.4 & 32.1 \\
		\hline
	    \multicolumn{4}{|c|}{Proposed} \\
		\hline
		sHCLM         & 22.0 & 26.8 & 31.8 \\
		sHCLM+Triplet & \bf 27.0 & \bf 30.4 & \bf 32.9 \\
		\hline
    \end{tabular} \\\ \\\ \\
	\begin{tabular}{|r|ccc|}
		\hline\rule{0pt}{1.1\normalbaselineskip}
		Dataset & \multicolumn{3}{c|}{ILSVRC-2012} \\
		\# Bits & 32 bits & 64 bits & 128 bits \\
		\hline
	    \multicolumn{4}{|c|}{Proxy embeddings} \\
		\hline
		Softmax-PQ~\cite{Sablayrolles:SHEval} & --- & 11.4 & --- \\
		BHC~\cite{Lin:BHC} & 10.5 & 14.2 & 17.4 \\
		\hline
	    \multicolumn{4}{|c|}{Triplet embeddings} \\
		\hline
		DTSH~\cite{Wang:TripletSH} & 9.3 & 11.6 & 13.5 \\
		\hline
	    \multicolumn{4}{|c|}{Proposed} \\
		\hline
		sHCLM & 10.3 & 14.4 & 17.4 \\
		sHCLM+Triplet & \bf 12.5 & \bf 16.5 & \bf 19.5 \\
		\hline
    \end{tabular}
\end{table}

\begin{table}[t!]
    \centering
    \caption{Classification performance (Acc) under transfer protocol on CIFAR-100 and ILSVRC-2012 datasets.}
    \label{tab:transfer-cls}
    \renewcommand{\arraystretch}{1.15}
	\begin{tabular}{|r|ccc|}
		\hline\rule{0pt}{1.1\normalbaselineskip}
		Dataset & \multicolumn{3}{c|}{CIFAR-100} \\
		Method & 16 bits & 32 bits & 64 bits \\
		\hline
	    \multicolumn{4}{|c|}{Classical} \\
		\hline
		LSH~\cite{Datar:LSH} & 31.6 & 41.3 & 47.9 \\
		ITQ~\cite{Gong:ITQ} & 42.7 & 51.6 & 56.3 \\
		KSH~\cite{Liu:KSH} & 38.3 & 45.5 & 47.8 \\
		SDH~\cite{Shen:SDH} & 35.6 & 42.9 & 46.2 \\
		ITQ-CCA~\cite{Gong:ITQ} & 37.6 & 44.4 & 50.6 \\
		\hline
	    \multicolumn{4}{|c|}{Proxy embeddings} \\
		\hline
		Softmax-PQ~\cite{Sablayrolles:SHEval} & --- & 47.4 & --- \\
		BHC~\cite{Lin:BHC} & 46.3 & 56.4 & 64.2 \\
		\hline
	    \multicolumn{4}{|c|}{Triplet embeddings} \\
		\hline
		DTSH~\cite{Wang:TripletSH} & 47.3 & \bf 58.0 & 64.4 \\
		\hline
	    \multicolumn{4}{|c|}{Proposed} \\
		\hline
        sHCLM & \bf 47.7 & 57.7 & \bf 65.0 \\
		sHCLM+Triplet & 44.9 & 55.6 & 63.6 \\
        \hline
	\end{tabular}\\\ \\\ \\ 
	\begin{tabular}{|r|ccc|}
		\hline\rule{0pt}{1.1\normalbaselineskip}
		Dataset & \multicolumn{3}{c|}{ILSVRC-2012} \\
		Method & 32 bits & 64 bits & 128 bits \\
		\hline
	    \multicolumn{4}{|c|}{Proxy embeddings} \\
		\hline
		BHC~\cite{Lin:BHC} & 32.8 & 42.1 & 50.1 \\
		\hline
	    \multicolumn{4}{|c|}{Triplet embeddings} \\
		\hline
		DTSH~\cite{Wang:TripletSH} & 30.9 & 40.3 & 48.2 \\
		\hline
	    \multicolumn{4}{|c|}{Proposed} \\
		\hline
        sHCLM & 35.4  & 46.1 & 54.5 \\
		sHCLM+Triplet & \bf 35.7 & \bf 46.3 & \bf 54.7 \\
        \hline
	\end{tabular}
\end{table}

\subsubsection{Comparison to prior work}
The transfer performance of the proposed embedding was compared 
to several approaches from the literature. 
\cite{Sablayrolles:SHEval} proposes an initial solution to the transfer 
setting, denoted Softmax-PQ, which learns a standard CNN classifier (AlexNet) 
and uses the product quantization (PQ) mechanism of~\cite{Jegou:PQ} to 
binarize the network softmax activations. Beyond this, we evaluated the 
transfer performance of the BHC and DTSH methods. Since these methods do 
not present results on unseen classes, we used the code released by the 
authors to train and evaluate each method under this protocol.

\paragraph{\it Retrieval:}
Table~\ref{tab:transfer-retr} shows the retrieval performance on 
CIFAR-100 and ILSVRC-2012. 
In the case of ILSVRC-2012, we trained AlexNet from scratch 
on the 750 training classes only, to ensure that images of test 
classes remain unseen until evaluation.
As expected, the retrieval mAP decreased significantly when compared 
to the non-transfer setting of Tables~\ref{tab:supervised-CIFAR100} 
and \ref{tab:supervised-ImageNet}. 
Similarly to the findings of Figure~\ref{fig:versions-transfer}, classical 
methods under-performed the more recent deep learning models. Also,
the triplet embedding DTSH outperforms the proxy embedding BHC and even 
sHCLM on CIFAR-100. 
However, the opposite occurs on ILSVRC-2012, where the number of classes 
is much larger. This is likely because, in this case, DTSH is not able to 
overcome the inefficiency of triplet sampling. 
Finally, while sHCLM overfits to the training classes, its combination 
with the triplet loss of \eqref{eq:tl}, sHCLM+Triplet, again achieves 
the best overall retrieval performance, on both datasets.

\paragraph{\it Classification:}
Table \ref{tab:transfer-cls} shows the performance of a classifier 
trained on the binary hash codes produced by each method. The conclusions 
are similar to those of Table~\ref{tab:transfer-retr}. The main 
difference is that the sHCLM proxy embedding has a classification 
performance much closer to that of the joint sHCLM+Triplet embedding, 
even outperforming the latter on CIFAR-100. This provides more evidence 
that proxy embeddings produce better local clusterings of the image 
embeddings and suggests that, when the goal is classification, there 
is little benefit in adding the triplet loss, even in the transfer setting.

\paragraph{\it Conclusions:} The transfer learning setting leads to a more
diverse set of conclusions than the standard supervised learning
setting. In fact, for transfer learning, the relative performances of different
approaches can vary substantially depending on whether the task is
classification or retrieval. For classification, the main conclusion from
the standard supervised setting continues to hold, i.e.,~there is very little
reason to consider any approach other than sHCML. However, for retrieval,
no clear winner emerges. Triplet embeddings can sometimes outperform
and sometimes underperform sHCML. Hence, in this setting, the combination of
sHCML and a triplet loss is beneficial. This combination achieves state-of-the-art retrieval performance in both datasets and has significant gains
over all other approaches in at least one dataset.

%%%%%%%%%%%%%%%%%%%%%%%%%%%%%%%%%%%%%%%%%%%%%%%%%%%%%%%%%%%%%%%%

\subsection{Learning to hash with multi-label similarities}

We finally evaluate the performance of hashing without explicit 
similarity classes. 
While this scenario can manifest itself in several ways, the most common 
example in the literature is the NUS-WIDE multi-tag dataset~\cite{NUS-WIDE}. 
%Since, in this dataset, similarity relations are not transitive, explicit 
%similarity classes cannot be formed directly. We define an approximate set 
%of classes, composed of the 334 most frequent tag combinations, which covers 
%90\% of all combinations present in the training images. All images in the 
%training set are then assigned to the class associated with the closest 
%tag combination. 
The network of~Figure~\ref{fig:deephash} is trained with an sHCLM proxy set, 
either using the binary cross-entropy loss alone (sHCLM) or the joint loss 
of~\eqref{eq:jointloss} (sHCLM+Triplet). The hyper-parameter $\lambda$ was 
tuned by cross-validation, using $\lambda \in \{0.01, 0.1, 1, 10, 100, 1000\}$.
Best performances were achieved for $\lambda =1$ or $\lambda = 10$,
depending on the number of bits of the hash code.

Table~\ref{tab:nus-wide} compares the retrieval performance of 
several methods. Note that the table differentiates the performance of
retrieval on the HBMP embedding with the standard Hamming distance
(denoted as HBMP bin) and with the floating point extension proposed
in~\cite{cakir2018hashing} (HBMP).
%Despite relying on an approximate set of
%similarity  classes, the
The sHCLM proxy embedding outperforms 
all ranking, triplet, and pair-wise embeddings that also use the Hamming
distance.
%and is competitive with the best triplet embedding. 
By combining sHCLM with a triplet loss, the proposed approach 
achieves the overall best performance, outperforming all methods that 
rely solely on binary operations for retrieval. The combination
of sHCLM+triplet and the Hamming distance even performs close to
HBMP~\cite{cakir2018hashing} which relies on floating-point operations 
for retrieval. Among strictly binary methods, the
only competitive approach is DSDH, which itself relies on a combination of a
proxy and a triplet loss.

\begin{table}[t!]
\centering
    \caption{Retrieval performance (mAP) on NUS-WIDE.\ \  * Performance obtained with retrieval procedures based on floating-point operations.}
    \label{tab:nus-wide}
    \begin{tabular}{|r|cccc|}
      \hline\rule{0pt}{1.1\normalbaselineskip}
      Method & 16 bits & 24 bits & 32 bits & 48 bits \\
      \hline
      \multicolumn{5}{|c|}{Ranking embeddings} \\
      \hline
      DSRH~\cite{Zhao:DSRH}& 60.9 & 61.8 & 62.1 & 63.1 \\
      \hline
      \multicolumn{5}{|c|}{Pair-wise embeddings} \\
      \hline
      DPSH~\cite{Li:DPSH}& 71.5 & 72.2 & 73.6 & 74.1 \\
      HBMP bin~\cite{cakir2018hashing} & 74.6 & -- & 75.4 & 75.4 \\
      HBMP~\cite{cakir2018hashing} & 80.4* & -- & 82.9* & 84.1* \\
      \hline
      \multicolumn{5}{|c|}{Triplet embeddings} \\
      \hline
      DRSCH~\cite{Zhang:BitScalable}& 61.8 & 62.2 & 62.3 & 62.8 \\
      DTSH~\cite{Wang:TripletSH}& 75.6 & 77.6 & 78.5 & 79.9 \\
      \hline
      \multicolumn{5}{|c|}{Combinations} \\
      \hline
      DSDH~\cite{Li:DeepSDH}& \bf 81.5 & 81.4 & 82.0 & 82.1 \\
      \hline
      \multicolumn{5}{|c|}{Proposed} \\
      \hline
      sHCLM & 79.3 & 80.2 & 80.2 & 80.9 \\
      sHCLM + Triplet & 81.4 & \bf 82.5 & \bf 83.0 & \bf 83.5 \\
      \hline
    \end{tabular}
\end{table}

\section{Conclusion}

In this work, we considered the hashing problem. We developed an
integrated understanding of classification and metric learning and have
shown that the rotational ambiguity of classification and retrieval losses
is a significant hurdle to the design of representations jointly optimal
for classification and hashing. We then proposed a new hashing procedure,
based on a set of fixed proxies, that eliminates this rotational ambiguity.
An algorithm was proposed to design semantic hash-consistent large margin
(sHCLM) proxies, which are nearly optimal for both classification and hashing.

An extensive experimental evaluation has provided evidence in support
of several important observations. First, sHCML was shown to
unequivocally advance the state-of-the-art in proxy-based hashing methods,
outperforming all previous methods in four datasets, two tasks (classification
and retrieval), and three hashing settings (supervised, transfer, and
multi-label). Second, for the setting where proxy embeddings are most popular,
namely supervised hashing, the gains were largest for the most challenging
datasets (CIFAR-100 and ILSVRC), where the number of classes is larger than
the dimension of the hashing code. For these datasets, sHCML, improved the
retrieval performance of the previous best proxy embeddings by as much as
10 points. To the best of our knowledge, no method in the literature has
comparable performance. Even the combination of sHCML with a triplet
embedding was unable to achieve consistent performance improvements. Third,
while proxy embeddings dominate in the classic supervised setting, this is less
clear for settings where class supervision is weaker, i.e. inference
has to be performed for classes unseen at training. This was the case
of both the transfer learning and multi-label datasets considered
in our experiments. While, in this setting, sHCML continued to dominate
for classification tasks, triplet embeddings were sometimes superior for
retrieval. Although no single method emerged as a winner for the retrieval task,
the combination of sHCML and a triplet loss was shown to achieve state-of-the-art  performance on all datasets considered.

Overall, sHCML achieved state-of-the-art results for all datasets, either
by itself (supervised setting or classification tasks) or when combined
with a triplet loss (retrieval tasks for settings with weak supervision).
These results show that it is an important contribution to the field
of proxy-based hashing embeddings. Nevertheless, they also show that
none of the two main current approaches to hashing, proxy and triplet
embeddings,
can fully solve the problem by itself. This suggests the need for
research on methods that can combine the best properties of each of
these approaches. More importantly, our results show that there is a
need to move beyond testing on a single hashing setting, a practice
that is still common in the literature. While we are not aware of any
previous work performing the now proposed joint evaluation over the
supervised, transfer, and multi-label settings, we believe that the wide
adaptation of this joint evaluation is critical for further advances in the
hashing literature.

%The resulting hashing procedure was shown to achieve state of the art results on several small and large scale datasets, for both classification and retrieval, both within and beyond the set of training classes.

\section*{Acknowledgments}
This work was funded by graduate fellowship 109135/2015 from the Portuguese Ministry of Sciences and Education, NSF Grants IIS-1546305, IIS-1637941, IIS-1924937, and NVIDIA GPU donations.

\bibliographystyle{spmpsci}
\bibliography{refs.bib}

\appendix

\section{Relations between classification and metric learning}
\label{app:equivalence}

Although seemingly different, metric learning and classification are closely related. 
To see this, consider the Bayes rule 
\begin{eqnarray}
  P_{Y|\X}(y|\x)
  &=&
  \frac{P_{\X|Y}(\x|y) P_Y(y)}{\sum_k P_{\X|Y}(\x|k) P_Y(k)}.
  \label{eq:bayes}
\end{eqnarray}
It follows from~(\ref{eq:softmax}) that
\begin{equation}
  P_{\X|Y}(\x|y) P_Y(y) \propto_\x  e^{\w^T_y \nu(\x) + b_y}
\end{equation}
where $\propto_\x$ denotes a proportional relation for each value of $\bf x$. This holds when
\begin{eqnarray}
  P_{\X|Y}(\x|y)
  & = &  q(\x) e^{\w^T_y \nu(\x) -
  \psi(\w_y)} \label{eq:expccd}\\
  P_Y(y) & =  & \frac{e^{b_y + \psi(\w_y)}}{\sum_k e^{b_k + \psi(\w_k)}},
  \label{eq:py}
\end{eqnarray}
where $q(\x)$ is any non-negative function and $\psi(\w_y)$ a constant such that~(\ref{eq:expccd}) integrates to one. In this case, $P_{\X|Y}(\x|y)$ is an exponential family distribution of canonical parameter $\w_y$, sufficient statistic $\nu(\x)$ and cumulant function $\psi(\w_y)$~\cite{barndorff2014information}. Further assuming, for simplicity, that the classes are balanced, i.e., $P_Y(y) =\frac{1}{C} \forall y$, leads to
\begin{eqnarray}
  b_y = -\psi(\w_y) + \log K
  \label{eq:uniform}
\end{eqnarray}
where $K$ is a constant.

The cumulant $\psi(\w_y)$ has several important properties~\cite{barndorff2014information,nelder1972generalized,banerjee2005clustering}. First, $\psi(\cdot)$ is a convex function of $\w_y$. Second, its first and second order derivatives are the mean $\nabla \psi(\w_y) = {\bf \mu}^\nu_y$ and co-variance $\nabla^2 \psi(\w_y) = {\bf \Sigma}^\nu_y$ of $\nu(\x)$ under class $y$. Third, $\psi(\cdot)$ has a conjugate function, convex on ${\bf \mu}^\nu_y$, given by 
\begin{equation}
  \phi({\bf \mu}^\nu_y) = \w^T_y {\bf \mu}^\nu_y -
  \psi(\w_y).
  \label{eq:conjugate}
\end{equation}
It follows that the exponent of~(\ref{eq:expccd}) can be re-written as
\begin{eqnarray}
  \w^T_y \nu(\x) -  \psi(\w_y)
  &=& \w^T_y {\bf \mu}^\nu_y - \psi(\w_y) +
      \w^T_y (\nu(\x) - {\bf \mu}^\nu_y) \nonumber \\
  &=& \phi({\bf \mu}^\nu_y) +
      \w^T_y (\nu(\x) - {\bf \mu}^\nu_y) \nonumber \\
  &=& \phi({\bf \mu}^\nu_y) +
        \nabla \phi({\bf \mu}^\nu_y)^T (\nu(\x) - {\bf \mu}^\nu_y) \nonumber \\
 &=& -d_{\phi}(\nu(\x),{\bf \mu}^\nu_y) + \phi(\nu(\x))
 \label{eq:connection}      
\end{eqnarray}
where
\begin{equation}
  d_{\phi}({\bf a},{\bf b}) =
  \phi({\bf a}) - \phi({\bf b}) - \langle\nabla \phi({\bf b}), {\bf a} - {\bf b} \rangle
\end{equation}
is the Bregman divergence between $\bf a$ and $\bf b$ associated with $\phi$. Thus,~(\ref{eq:expccd}) can be written as
\begin{equation}
  P_{\X|Y}(\x|y)
  =  u(\x) e^{-d_{\phi}(\nu(\x),{\bf \mu}^\nu_y)}
\end{equation}
where $u(\x) = q(\x) e^{\phi(\nu(\x))}$ and using (\ref{eq:uniform}), (\ref{eq:py}) and (\ref{eq:bayes}), 
\begin{eqnarray}
  P_{Y|\X}(y|\x)
  &=& \frac{e^{-d_{\phi}(\nu(\x),{\bf \mu}^\nu_y)}}
      {\sum_{k} e^{-d_{\phi}(\nu(\x),{\bf \mu}^\nu_k)}}.
      \label{eq:hy2}
\end{eqnarray}
Hence, learning the embedding $\nu(\x)$ with the softmax classifier of~(\ref{eq:softmax}) endows $\cal V$ with the Bregman divergence $d_{\phi}(\nu(\x),{\bf \mu}^\nu_y)$.
From (\ref{eq:conjugate}), it follows that
\begin{equation}
  \nabla \psi(\w_y) = {\bf \mu}_y^\nu  \quad \quad \quad
  \nabla \phi({\bf \mu}_y^\nu) = \w_y.
\end{equation}
Hence,
\begin{equation}
  {\bf \mu}_y^\nu = \w_y
  \label{eq:mu=w}
\end{equation}
if and only if
\begin{eqnarray}
  \nabla \psi(\w_y) = \w_y  \quad \quad \quad
  \nabla \phi({\bf \mu}_y^\nu) = {\bf \mu}_y^\nu,
\end{eqnarray}
which holds when
\begin{eqnarray}
  \psi({\bf a}) = \phi({\bf a}) = \frac{1}{2}||{\bf a}||^2.
  \label{eq:psi}
\end{eqnarray}
It can be shown that the corresponding exponential family model is the Gaussian of identity covariance and the corresponding Bregman divergence the squared Euclidean distance. Hence, ${\bf \mu}_y^g = \w_y$ if only if $d_\phi$ is the $L_2$ distance. In this case, (\ref{eq:hy2}) reduces to
\begin{eqnarray}
  P_{Y|\X}(y|\x)
  &=& \frac{e^{-d(\nu(\x),\w_y)}}
      {\sum_{k} e^{-d(\nu(\x),\w_k)}}.
\end{eqnarray}

\end{document}